\documentclass[10pt,journal]{IEEEtran}
\usepackage[pdftex]{graphicx}
\DeclareGraphicsExtensions{.pdf,.jpg,.png}
\usepackage{cite}
\usepackage[cmex10]{amsmath}
\usepackage{amssymb}
\usepackage{graphicx}
\usepackage{cases}
\usepackage{mdwmath}
\usepackage{mdwtab}
\usepackage{array}
\usepackage{url}
\usepackage[ruled,vlined]{algorithm2e}
\usepackage{booktabs}
\usepackage{threeparttable}
\usepackage{multirow}
\usepackage{color}
\usepackage{caption}
\def\ie{\emph{i.e.}}
\def\eg{\emph{e.g.}}
\def\etal{{\em et al.}}

\newcommand{\myPara}[1]{\vspace{.05in}\noindent\textbf{#1}}
\newcommand{\rev}[1]{\textcolor{black}{#1}}
\newcommand{\rr}[1]{\textcolor{red}{#1}}
\newcommand{\cg}[1]{\textcolor{green}{#1}}
\newcommand{\bb}[1]{\textcolor{blue}{#1}}
\newcommand{\bl}[1]{\textbf{#1}}

\newcommand{\mc}[1]{\mathcal{#1}}
\newcommand{\mb}[1]{\mathbb{#1}}

\graphicspath{{./figs/}}
\usepackage{color}
\begin{document}

\title{A Benchmark Dataset and Saliency-guided Stacked Autoencoders for Video-based Salient Object Detection}
\author{Jia~Li,~\IEEEmembership{Senior Member,~IEEE},~Changqun Xia~and~Xiaowu~Chen,~\IEEEmembership{Senior Member,~IEEE} %  % <-this % stops a space
\IEEEcompsocitemizethanks{
\IEEEcompsocthanksitem J. Li, C. Xia and X. Chen are with the State Key Laboratory of Virtual Reality Technology and Systems, School of Computer Science and Engineering, Beihang University, Beijing, 100191, China. \protect
\IEEEcompsocthanksitem J. Li is also with the International Research Institute for Multidisciplinary Science at Beihang University, Beijing, 100191, China. \protect
%\IEEEcompsocthanksitem Correspondence should be addressed to Jia Li and Xiaowu Chen. Email: \{jiali, chen\}@buaa.edu.cn.\protect
%\IEEEcompsocthanksitem Manuscript received 2016.
%funding support goes here
}}

%\markboth{SUBMISSION TO IEEE TRANSACTIONS ON IMAGE PROCESSING}%
%{Li \MakeLowercase{\textit{{\em et al.}}}: Video-based Salient Object Detection}

% PRIOR to the title within the \IEEEcompsoctitleabstractindextext IEEEtran
\IEEEcompsoctitleabstractindextext{%
\begin{abstract}
Image-based salient object detection (SOD) has been extensively studied in the past decades. However, video-based SOD is much less explored since there lack large-scale video datasets within which salient objects are unambiguously defined and annotated. Toward this end, this paper proposes a video-based SOD dataset that consists of 200 videos (64 minutes). In constructing the dataset, we manually annotate all objects and regions over 7,650 uniformly sampled keyframes and collect the eye-tracking data of 23 subjects that free-view all videos. From the user data, we find salient objects in video can be defined as objects that consistently pop-out throughout the video, and objects with such attributes can be unambiguously annotated by combining manually annotated object/region masks with eye-tracking data of multiple subjects. To the best of our knowledge, it is currently the largest dataset for video-based salient object detection.

Based on this dataset, this paper proposes an unsupervised baseline approach for video-based SOD by using saliency-guided stacked autoencoders. In the proposed approach, multiple spatiotemporal saliency cues are first extracted at pixel, superpixel and object levels. With these saliency cues, stacked autoencoders are unsupervisedly constructed which automatically infer a saliency score for each pixel by progressively encoding the high-dimensional saliency cues gathered from the pixel and its spatiotemporal neighbors. \rev{Experimental results show that the proposed unsupervised approach outperforms 30 state-of-the-art models on the proposed dataset, including 19 image-based \& classic (unsupervised or non-deep learning), 6 image-based \& deep learning, and 5 video-based \& unsupervised.} Moreover, benchmarking results show that the proposed dataset is very challenging and has the potential to boost the development of video-based SOD.

\end{abstract}
% not want either math or citations to appear in the abstract.

% Note that keywords are not normally used for peer review papers.
\begin{keywords}
Salient object detection, video dataset, stacked autoencoders, model benchmarking
\end{keywords}}

\maketitle

% is not selected <OR> if conference mode is selected - because compsoc
% conference papers position the abstract like regular (non-compsoc)
\IEEEdisplaynotcompsoctitleabstractindextext

% \fi
\IEEEpeerreviewmaketitle

\section{Introduction}
\IEEEPARstart{T}{he} booming of image-based salient object detection (SOD) originates from the presence of large-scale benchmark datasets \cite{LiuSZTS07Learn,achanta2009frequency}. With these datasets, it becomes feasible to construct complex models with machine learning algorithms (\eg, random forest regressor~\cite{JiangWYWZL13}, bootstrap learning~\cite{Tong15bootstrap}, multi-instance learning~\cite{han2016co} and deep learning~\cite{Zhao15Deep}). Moreover, the presence of such datasets enables fair comparisons between state-of-the-art models~\cite{borji2012salient,SalObjBenchmark}. Actually, large-scale datasets provide a solid foundation for SOD and consistently guide the development of this area.

In the past decade, SOD datasets keep on evolving to meet the increasing demands in developing and benchmarking new models. Some researchers argue that images in early datasets like \bl{ASD}~\cite{achanta2009frequency} and \bl{MSRA-B}~\cite{LiuSZTS07Learn} are relatively small and simple. They extend such datasets in terms of amount \cite{MSRA10KTHUR15Kdb,YangZLRY13Manifold} or complexity \cite{yan2013hierarchical,borjiTIP2014,liXiaodiCVPR2014}. Meanwhile, the concept of SOD has been extended to RGBD images~\cite{Peng14rgbd}, image collections~\cite{li2013co,Fu13Cluster,Zhang15MIL} and videos \cite{rahtu2010segmenting,li2013exploring,Fu15Video,wang2015GF}. Among these extensions, video-based SOD has invoked great research interests since it re-defines the problem from a spatiotemporal perspective. However, there still lack large-scale video datasets for comprehensive model comparison, which prevents the fast growth of this branch. For example, the widely used \bl{SegTrack} dataset~\cite{tsai2010segtrack} consists of only 6 videos with 21 to 71 frames per video, while a recent dataset \bl{ViSal}~\cite{wang2015GF} contains only 17 videos with 30 to 100 frames per video. In addition, the definition of salient object in video is still not very clear (\eg, manually annotated foreground objects~\cite{perazzi2016davis}, class-specific objects \cite{wang2015GF} or moving objects \cite{brox2010longterm}). It is necessary to construct a large video dataset with unambiguously defined salient objects.

\begin{figure*}[t]
\centering
\includegraphics[width=1.00\textwidth]{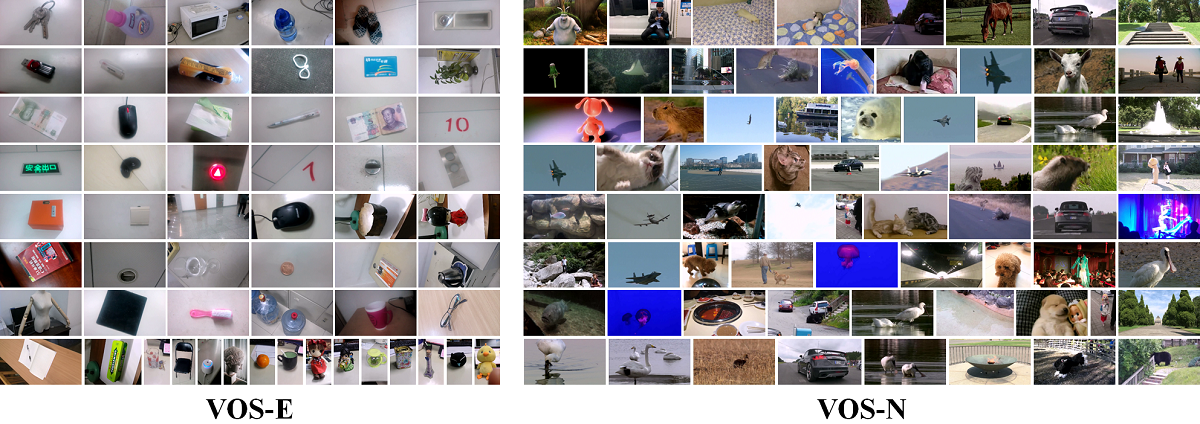}
\caption{Representative scenarios in \bl{VOS}. The 200 videos in \bl{VOS} are grouped into two subsets according to the complexity of foreground, background and motion, including \bl{VOS-E} (\textit{easy} subset, 97 videos) and \bl{VOS-N} (\textit{normal} subset, 103 videos). }
\label{fig:representativeVideo}
\end{figure*}

To address this issue, this paper proposes \bl{VOS}, \rev{a large-scale dataset with 200 indoor/outdoor videos for video-based SOD} (64 minutes, $116,103$ frames, see Fig.~\ref{fig:representativeVideo} for representative scenarios). In constructing \bl{VOS}, we first collect two types of user data, including 1)~the eye-tracking data of 23 subjects that free-view all the 200 videos, and 2)~the masks of all objects and regions in $7,650$ uniformly sampled keyframes annotated by another 4 subjects. Based on these user data, salient objects in a video can be unambiguously annotated as the objects that consistently receive the highest density of fixations throughout the video. After discarding the pure-background keyframes as well as the keyframes in which salient objects are partially occluded or split into several disjoint parts, we obtain $7,467$ keyframes with binary masks of salient objects.

\rev{Based on the large-scale dataset, we propose an unsupervised baseline model for video-based SOD by constructing saliency-guided stacked autoencoders. Different from the fixation prediction task that aims to roughly detect where the human-being looks at and the image-based SOD task that aims to segment only the most spatially salient objects, the video-based SOD focuses on detecting and segmenting the objects that consistently pop-out throughout a video from a spatiotemporal perspective. Inspired by this fact, the proposed approach first extracts multiple spatiotemporal saliency cues at pixel, superpixel and object levels}. Stacked autoencoders are then unsupervisedly trained which can automatically infer a saliency score for each pixel by progressively encoding the high-dimensional saliency cues gathered from the pixel and its spatiotemporal neighbors. \rev{In the comprehensive model benchmarking on \bl{VOS}, the proposed approach outperforms 30 image-based and video-based models. Moreover, the benchmarking results validate that \bl{VOS} is a challenging dataset that has the potential to greatly boost the development of this area.}

Our main contributions are summarized as follows: \rev{1)~we propose a large and challenging dataset for video-based SOD, which we believe can be useful for the development of this area, 2)~we propose saliency-guided stacked autoencoders for video-based SOD, which is an unsupervised baseline model that outperforms 30 image-based and video-based models, and 3) we provide a comprehensive benchmark of our approach and massive \rev{state-of-the-art models}, which reveals several key challenges in video-based SOD and further validates the usefulness of the proposed dataset.}

The rest of this paper is organized as follows: Section~\uppercase\expandafter{\romannumeral2} reviews existing datasets and models. Section~\uppercase\expandafter{\romannumeral3} presents a new dataset. In Section~\uppercase\expandafter{\romannumeral4}, we propose saliency-guided stacked autoencoders for video-based SOD. Section~\uppercase\expandafter{\romannumeral5} benchmarks the proposed model and the state-of-the-art models, and the paper is concluded in Section~\uppercase\expandafter{\romannumeral6}.

% the motion of foreground object may not always differs from the background. In video, everything may happen.

\section{Related Work}
Video-based SOD is correlated with image-based SOD, foreground/primary object detection and moving object segmentation. In this section, we will review the most related datasets and models from all these areas.

\begin{table*}[t]
\scriptsize
\centering{
\caption{\rev{Comparison between \bl{VOS} (subsets: \bl{VOS-E} and \bl{VOS-N}) with representative image/video object segmentation datasets}}
\label{tab:datasetComparison}
\begin{tabular}{l@{}c|c|ccc|cc|cc|c|c} \hline
\multicolumn{2}{c|}{\multirow{2}*{Dataset}}        & \multirow{2}*{\#Vid.} & \multicolumn{3}{c|}{Resolution (in pixels)} &\multicolumn{2}{c|}{\#Orig. Frames} & \multicolumn{2}{c|}{\#Labeled Frames} & \#Avg. Obj$^*$. & Obj. Area   \\
           &                     &                 & Width  & Height  & Max Res.  & Total & Avg.          & Total & Avg.        & Per Frame &  Per Frame (\%)  \\ \hline
\bl{\multirow{8}{*}{\rotatebox{90}{Image-based}}}
&~\bl{ASD}~\cite{achanta2009frequency}     & -     & $[188,400]$  & $[165,400]$   & $400\times{}400$  & 1000 & -            & 1000  & -         & 1.16$\pm$0.87 & $19.9\pm9.52$\\
&~\bl{ECSSD}~\cite{yan2013hierarchical}    & -     & $[222,400]$  & $[139,400]$   & $400\times{}400$  & 1000 & -            & 1000  & -         & 1.16$\pm$0.56 & $28.5\pm20.5$\\
&~\bl{DUT-O}~\cite{YangZLRY13Manifold}     & -     & $[167,401]$  & $[89,401]$    & $401\times{}401$  & 5168 & -            & 5168  & -         & 1.20$\pm$0.69 & $14.9\pm12.2$\\
&~\bl{PASCAL-S}~\cite{liXiaodiCVPR2014}    & -     & $[191,500]$  & $[151,500]$   & $500\times{}500$  & 850  & -            & 850   & -         & 1.52$\pm$1.11 & $24.7\pm16.0$\\
&~\bl{MSRA10K}~\cite{MSRA10KTHUR15Kdb}     & -     & $[179,400]$  & $[165,400]$   & $400\times{}400$  & 10000& -            & 10000 & -         & 1.05$\pm$0.46 & $22.2\pm10.1$\\
&~\bl{HKU-IS}~\cite{li2015mdf}             & -     & $[100,401]$  & $[100,401]$   & $400\times{}400$  & 4447 & -            & 4447  & -         & 1.60$\pm$0.82 & $19.1\pm10.9$\\
%&~\bl{THUR15K}~\cite{MSRA10KTHUR15Kdb}     & -    & $[300,1208]$ & $[300,828]$   & $1208\times{}300$ & 6233 & -            & 6233  & -         & 1.02$\pm$0.19 & $15.1\pm12.5$\\
&~\bl{XPIE}~\cite{xia2017what}             & -     & $[155,500]$  & $[130,500]$   & $500\times{}500$  & 10000& -            & 10000 & -         & 1.16$\pm$0.46 & $19.4\pm14.4$\\
\hline
\bl{\multirow{8}{*}{\rotatebox{90}{Video-based}}}
&~\bl{SegTrack}~\cite{tsai2010segtrack}    &  6    & $[259,414]$  & $[212,352]$   & $414\times{}352$  & 244  & 41$\pm$18    & 244   & 41$\pm$18 & 1.00$\pm$0.00 & $3.46\pm2.84$\\
&~\bl{SegTrack V2}~\cite{li2013segtrackv2} & 14    & $[259,640]$  & $[212,360]$   & $640\times{}360$  & 1065 & 76$\pm$82    & 1065  & 76$\pm$82 & 1.38$\pm$1.01 & $7.38\pm7.89$\\
&~\bl{FBMS}~\cite{ochs2014moseg}            & 59    & $[350,960]$  & $[253,540]$   & $960\times{}540$  & 13860& 235$\pm$193  & 720   & 12$\pm$8  & 1.78$\pm$1.54 & $14.4\pm13.7$\\
&~\bl{DAVIS}~\cite{perazzi2016davis}       & 50    & $[1600,1920]$& $[900,1080]$  & $1920\times{}1080$& 3455 & 69$\pm$19    & 3455  & 69$\pm$19 & 5.39$\pm$22.87& $8.10\pm6.44$\\
&~\bl{ViSal}~\cite{wang2015GF}             & 17    & $[320,512]$  & $[240,288]$   & $512\times{}288$  & 963  & 57$\pm$20    & 193   & 11$\pm$4  & 1.16$\pm$0.40 & $10.5\pm6.51$\\
&~\bl{VOS-E}                               & 97    & $[408,800]$  & $[448,800]$   & $800\times{}640$  & 49206& 507$\pm$130  & 3236  & 33$\pm$9  & 1.02$\pm$0.18 & $18.4\pm12.8$\\
&~\bl{VOS-N}                               & 103   & $[448,800]$  & $[312,800]$   & $800\times{}800$  & 66897& 649$\pm$510  & 4231  & 41$\pm$33 & 1.25$\pm$0.54 & $8.92\pm10.8$\\
&~\bl{VOS}                                 & 200   & $[408,800]$  & $[312,800]$   & $800\times{}800$  &116103& 581$\pm$383  & 7467  & 37$\pm$25 & 1.15$\pm$0.44 & $13.0\pm12.6$\\ \hline
\end{tabular}
\begin{tablenotes}
\footnotesize
\item[1] $^*$ Objects are counted as disconnected foreground regions. In \bl{DAVIS}, a semantic object may be divided into hundreds of disconnected parts (\eg, a bus occluded by a tree), leading to a extremely high mean and standard deviation in the number of foreground ``objects'' per frame.
\end{tablenotes}
}
\end{table*}

\subsection{Datasets}

\myPara{SegTrack}~\cite{tsai2010segtrack} is a popular dataset for video object segmentation. It contains 6 videos about animal and human with 244 frames in total, and videos are intentionally collected for benchmarking models with predefined challenges. Only one foreground object is manually annotated per frame.

\myPara{SegTrack V2}~\cite{li2013segtrackv2} extends \bl{SegTrack} from two perspectives. First, additional annotations of foreground objects are provided for the six videos in \bl{SegTrack}. Second, 8 new videos are carefully chosen to cover more challenges. In total, \bl{SegTrack V2} contains 14 videos about bird, animal, car and human with $1,065$ densely annotated frames.

\myPara{Freiburg-Berkeley Motion Segmentation (FBMS)} is designed for motion segmentation (\ie, segmenting regions with similar motion). It is first proposed in \cite{brox2010longterm} with 26 videos, and then Ochs~\etal\cite{ochs2014moseg} extended the dataset with another 33 videos. In total, this dataset contains 59 videos with 720 sparsely annotated frames. Although the dataset is much larger than \bl{SegTrack} and \bl{SegTrack V2}, the scenarios it covers are still far from sufficient~\cite{perazzi2016davis}. Moreover, moving objects are not equivalent to salient objects, especially in a scene with complex content.

\myPara{DAVIS}~\cite{perazzi2016davis} contains 50 high quality videos about human, animal, vehicle, object and action with $3,455$ densely annotated frames. Each video has Full HD 1080p resolution and lasts about 2 to 4 seconds. \rev{Each video clip in this dataset contains one foreground object or two spatially connected objects. Note that such objects may split into hundreds of small regions due to occlusion.}

\myPara{ViSal} is a pioneer video-based SOD dataset proposed in \cite{wang2015GF}. It contains 17 videos about human, animal, motorbike, etc. Each video contains 30 to 100 frames, in which salient objects are manually annotated according to the semantic classes of videos. In other words, this dataset assumes that salient objects are equivalent to the primary objects within videos annotated by semantic tags.

\rev{To facilitate the comparison between these datasets and our \bl{VOS} dataset, we show in Table~\ref{tab:datasetComparison} more dataset statistics. Moreover, we also demonstrate the details of 7 representative image-based SOD datasets so as to provide an intuitive impression between image- and video-based SOD.} Generally speaking, previous datasets reviewed above have greatly boosted the researches in video object segmentation but still have several drawbacks.

First, these datasets are still a little small for modern learning algorithms like Convolutional Neural Networks (CNN). \rev{As shown in Table~\ref{tab:datasetComparison}, the numbers of annotated frames in most previous video datasets are much smaller than the image-based SOD datasets and \bl{VOS}.} Although thousands of frames in \bl{SegTrack V2} and \bl{DAVIS} are densely annotated, the rich redundancy in consecutive frames may increase the over-fitting risk in model training.

\rev{Second, videos in some datasets are selected to maximally cover predefined challenges in video object segmentation~(\eg, \bl{SegTrack} and \bl{SegTrack V2}). However, such intentionally selected videos may make the dataset not very ``realistic'' (\ie, different from the videos in real-world scenarios). Moreover, such datasets may favor models that are particularly designed to ``over-fit'' the limited scenarios. On the contrary, our \bl{VOS} dataset is much larger so that the over-fitting risk can be largely alleviated}.

Third, foreground objects in previous datasets are often manually annotated only by one or several annotators, which may incorporate strong subjective bias into these datasets. For example, in a video with both dog and monkey only the monkey is annotated in \bl{SegTrack}, while \bl{SegTrack V2} has the dog annotated as well. Actually, manual annotations from different subjects often conflict with each other~\cite{liu2011learning} and cause ambiguity. \rev{To alleviate such ambiguity, previous works like \cite{li2009dataset,carmi2006role,vigier2016uheddataset} have tried to locate salient targets by averaging rectangles manually annotated by 23 subjects~\cite{li2009dataset} or collecting human fixations via eye-tracking apparatus~\cite{carmi2006role,vigier2016uheddataset}. However, these datasets cannot be directly used in video-based SOD for lacking pixel-wise annotations of salient objects. Actually, pixel-wise annotation is the most time-consuming procedure in constructing video-based SOD datasets like \bl{VOS}.}

To sum up, existing datasets are still a little insufficient to benchmark video-based SOD models due to the limited video numbers as well as the ambiguous definition and annotation processes of salient/foreground/moving objects. To further boost the development of this area, it is necessary to construct a large-scale dataset that covers a wide variety of real-world scenarios and contains salient objects that are unambiguously defined and annotated.

\subsection{Models}
\rev{Hundreds of bottom-up and learning-based models~\cite{ChengCVPR,shen2012unified,JiangWYWZL13,Jiang2013SaliencyMC,zhu2014saliency,Zhang2015MB} have been proposed for image-based SOD in the past decade. With the booming of deep learning methodology and the presence of large-scale datasets~\cite{li2015mdf,yang2013manifold,ChengPAMI}, many deep models~\cite{wang2015legs,li2016dcl,zhao2015mcdl,hou2017ehed} have been proposed for image-based SOD. For example, Han~\etal~\cite{han2015drr} proposed multi-stream stacked denoising autoencoders that can detect salient regions by measuring the reconstruction residuals that reflect the distinctness between background and salient regions. He~\etal~\cite{he2015supercnn} adopted CNNs to characterize superpixels with hierarchical features so as to detect salient objects at multiple scales, while the superpixel-based saliency computation was used by \cite{lee2016eld,li2015mdf} as well. Considering that the task of fixation prediction is tightly correlated with SOD, a unified deep network was proposed in \cite{srinivas2016su} for simultaneous fixation prediction and image-based SOD.}

\rev{The state-of-the-art deep SOD models often adopt recurrent frameworks that can achieve impressive performance. For example, Liu~\etal~\cite{liu2016dhsnet} adopted hierarchical recurrent CNNs to progressively refine the details of salient objects. In \cite{jason2016recurrent}, a coarse saliency map was first generated by using the convolution-deconvolution networks. After that, it was refined by iteratively enhancing the results in various sub-regions. Wang~\etal~\cite{wang2016rfcn} iteratively delivered the intermediate predictions back to the recurrent CNNs to refine saliency maps. In this way, salient objects can gradually pop-out, while distractors can be progressively suppressed.}

Compared with image-based SOD, video-based SOD is less explored due to the lack of large video datasets. For example, Liu~\etal~\cite{liu2008video} extended their image-based SOD model \cite{LiuSZTS07Learn} to the spatiotemporal domain for salient object sequence detection. In \cite{fukuchi2009saliencyseg}, visual attention (\ie, the estimated fixation density) was used as prior knowledge to guide the segmentation of salient regions in video. Rahtu~\etal~\cite{rahtu2010segmenting} proposed to integrate local contrast features in illumination, color and motion channels with a statistical framework. A conditional random field was then adopted to recover salient objects from images and video frames. Due to the lack of large-scale benchmarking datasets, most of these early approaches only provide qualitative comparisons, and only a few works like \cite{liu2008video} have provided quantitative comparisons on a small dataset within which salient objects are roughly annotated with rectangles.

To conduct quantitative comparisons in single video-based SOD, Bin~\etal~\cite{bin2013temporally} manually annotated the salient objects in 10 videos with about 100 frames per video. They also proposed an approach to detect temporally coherent salient objects using regional dynamic contrast features in the spatiotemporal domain of color, texture and motion. Their approach demonstrated impressive performance in processing videos with only one salient object. In \cite{papazoglou2013fst}, Papazoglou and Ferrari proposed an approach for the fast segmentation of foreground objects from background regions. They first estimated an initial foreground map with respect to the motion information, which was then refined by building the foreground/background appearance models and encouraging the spatiotemporal smoothness of foreground objects over the whole video. The main assumption required by their approach was that foreground objects should move differently from its surrounding background in a good fraction of the video. Wang~\etal~\cite{wang2015sag} proposed an unsupervised approach for video-based SOD. In their approach, frame-wise saliency maps were first generated and refined with respect to the geodesic distances between regions in the current frame and subsequent frames. After that, global appearance models and dynamic location models were constructed so that the spatially and temporally coherent salient objects can be segmented by using an energy minimization framework. In their later work~\cite{wang2015GF}, Wang~\etal~proposed to utilize the inter-frame and intra-frame information in a gradient flow field. By extracting the local and global saliency measures, an energy function was then adopted to enhance the spatiotemporal consistency of the output saliency maps.

Despite the performance and benchmarking methodologies, these single video-based approaches have provide us an intuitive definition of salient objects. That is, salient objects in a video should be spatiotemporally consistent and visually distinct from background regions. However, in real-world scenarios the assumptions like color/texture dissimilarity and motion irregularity may not always hold. A more general definition of salient objects in video is required to guide the annotation and detection processes.

Beyond single video-based approaches, some approaches extend the idea of image co-segmentation to the video domain. For example, Chiu and Fritz~\cite{chiu2013coseg} proposed a generative model for multi-class video co-segmentation. A global appearance model was learned to connect the segments from the same class so as to segment the foreground targets shared by different videos. Fu~\etal~\cite{Fu15Video} proposed to detect multiple foreground objects shared by a set of videos. Category-independent object proposals were first extracted and multi-state selection graph was then adopted to handle multiple foreground objects. Although video co-segmentation brings us a interesting new direction for studying video-based SOD, detecting salient objects in a single video is still the most common requirement in many real-world applications.

\section{A Large-scale Dataset for Video-based SOD}

A good benchmark dataset should cover many real-world scenarios and the annotation process should contain little subjective bias. In this section, we will introduce the details in constructing the dataset and discuss how salient objects can be unambiguously defined and annotated in videos.

%Dataset: Large, realistic, and less confusing.

\subsection{Video Collection}
We first collect hundreds of long videos from Internet (\eg, video-sharing websites like Youtube) and volunteers. Note that no instruction is given on what types of videos are required since we aim to collect more ``realistic'' daily videos. After that, we randomly sample short clips from long videos and keep only the clips that contain objects in most frames. Finally, we obtain 200 indoor/outdoor videos that last 64 minutes in total ($116,103$ frames at 30fps). These videos are grouped into two subsets according to the content complexity, including:

\myPara{VOS-E}. This subset contains 97 \textit{easy} videos (27 minutes, $49,206$ frames, 83 to 962 frames per video). As shown in Fig.~\ref{fig:representativeVideo}, a video in this subset usually contains obvious foreground objects with slow camera motion. This subset serves as a baseline to explore the inherent correlations between image- and video-based SOD.

\myPara{VOS-N}. This subset contains 103 \textit{normal} videos (37 minutes, $66,897$ frames, 710 to $2,249$ frames per video). As shown in Fig.~\ref{fig:representativeVideo}, videos in this subset contain complex or highly dynamic foreground objects, dynamic or cluttered background regions, etc. This subset is very challenging and can be used to benchmark models in realistic scenarios.

\subsection{User Data Collection}
The manual annotation of salient objects often generate ambiguity and strong subjective bias in complex scenes. Inspired by the solution used in \cite{liXiaodiCVPR2014}, we collect two types of user data, including object masks and human fixations, to alleviate the ambiguity in defining and annotating salient objects in videos.

\begin{figure}[t]
\centering
\includegraphics[width=1.0\columnwidth]{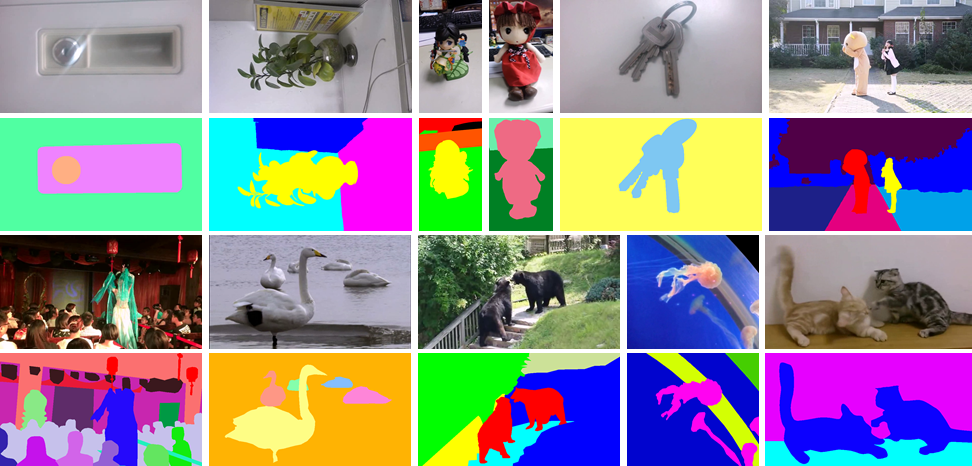}
\caption{Masks of objects and regions annotated by 4 subjects. Holes are filled up to speed up the annotation process (\eg, the key in the first row), and multiple objects will be assigned the same labels throughout the video if they cannot be easily separated in certain frames (\eg, the fighting bears and cats) or difficult to be re-identified (\eg, the jelly fishes which frequently appear and disappear near screen borders).}
\label{fig:objectMasks}
\end{figure}

\myPara{Object masks}. Four subjects (2 males and 2 females, aged between 24 and 34) manually annotate the accurate boundaries of all objects and regions in video frames. Since it consumes too much time to annotate all frames, we uniformly sample only one keyframe out of every 15 frames and manually annotate the $7,650$ keyframes. In the annotation, an object will maintain the same label throughout a video, and the holes in objects are filled to speed up the annotation. Since moving objects may merge or split several times in a short period and it is difficult to consistently assign different labels to them (\eg, the fighting bears and cats in the third row of Fig.~\ref{fig:objectMasks}), we assign the same label to objects if they become indistinguishable in certain frames (\eg, the bears and cats in Fig.~\ref{fig:objectMasks}) or difficult to be re-identified (\eg, the jelly fishes in Fig.~\ref{fig:objectMasks} frequently appear and disappear near screen borders). Finally, regions smaller than 16 pixels are ignored and we obtain the accurate boundaries of $53,478$ objects and regions.

\myPara{Human fixations}. Twenty-three subjects (16 males and 7 females, aged between 21 and 29) participate in the eye-tracking experiments. Note that none of them participates in annotating the object/region masks. Each subject is asked to free-view all the 200 videos displayed on a 22-inch color monitor with a resolution of $1680\times{}1050$. A chin rest is adopted to reduce head movements and enforce a viewing distance of $75$cm. \rev{Considering that the non-stop watching of 200 videos (64 minutes) will be very tiring, we randomly divide videos into subgroups and adopt an interlaced schedule for different subjects that free-view the same subgroup of videos. In this manner, each subject will get sufficient time to rest after watching a small collection of videos, making the eye-tracking data more reliable.} During the free-viewing process, an eye-tracking apparatus with a sample rate of 500Hz \rev{(SMI RED 500)} is used to record various types of eye movements. Finally, we keep only the \textit{fixations} and denote the set of \rev{eye positions} on a video $\mathcal{V}$ as $\mathbb{F}_\mathcal{V}$, in which \rev{a sampled eye position} $f$ is represented by a triplet $(x_f,y_f,t_f)$. Note that $x_f$ and $y_f$ are the coordinates of $f$ and $t_f$ is the time stamp that $f$ starts (\rev{an eye position sampled by the 500HZ eye-tracker lasts about two milliseconds}, see Fig.~\ref{fig:badFixation} for some examples).

\subsection{Definition and Annotation of Salient Objects in Video}
In early datasets with only simple images, salient objects can be manually annotated without much ambiguity. However, in a complex video there may exist several candidate objects, and different subjects may have different biases in determining which ones are the most salient. As a result, such subjective biases prevent the direct manual annotation of salient objects in complex videos.

\begin{figure}[t]
\centering
\includegraphics[width=1.00\columnwidth]{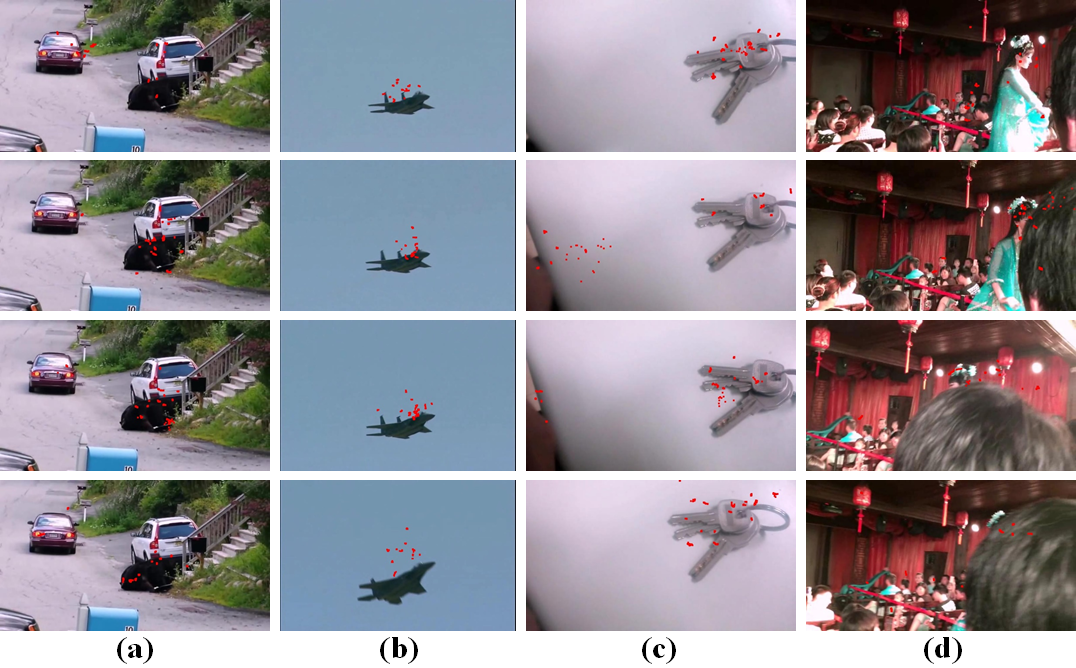}
\caption{Human fixations (red dots) of 23 subjects on consecutive keyframes. We can see that these fixations are insufficient to directly annotate salient objects frame by frame. (a)~Insufficient fixations to distinguish multiple salient objects and distractors; (b)~Fixations fall outside small moving objects; (c)~Fixations distracted by visually surprising regions; (d)~Salient objects occluded by background regions, leading to background-only frames.}
\label{fig:badFixation}
\end{figure}

To alleviate the subjective bias, the fixations of multiple subjects can be used to find the most salient objects. For example, Li~\etal~\cite{liXiaodiCVPR2014} collect fixations from 8 subjects that free-view the same image for 2 seconds. After that, salient objects are defined as the objects that receive the highest number of fixations. This solution provides a less ambiguous definition of salient objects in images but may fail on videos due to four reasons:

\myPara{1)~Insufficient viewing time}. The viewing time of a frame (\eg, 33ms) is much shorter than that of an image. As a result, the fixations received by a frame are often insufficient to fully distinguish the most salient objects, especially when there exist multiple candidates in the same video frame (\eg, the cars and bears in Fig.~\ref{fig:badFixation} (a)).

\myPara{2)~Inaccurate fixations}. Human fixations may fall outside moving objects and small objects (\eg, the fast moving aircraft in Fig.~\ref{fig:badFixation} (b)).

\myPara{3)~Rapid attention shift}. Human attention can be suddenly distracted by visual surprise and then return to the salient objects after a short period. In this case, the surprising background regions will be mistakenly recognized as salient if only the fixations in this short period are considered in defining salient objects (\eg, the black region in Fig.~\ref{fig:badFixation} (c)).

\myPara{4)~Background-only frames}. Some frames are purely background. If salient objects are defined by fixations received only by these frames, background regions in these frames will be mistakenly annotated as salient (\eg, the girl is occluded by background regions in Fig.~\ref{fig:badFixation} (d)).

\rev{For these reasons}, it is difficult to directly define and annotate salient objects separately on each frame. Inspired by the idea of co-saliency~\cite{chiu2013coseg,Fu15Video}, we propose to define salient objects at the scale of whole videos. That is, salient objects in videos are defined as \textit{the objects that consistently receive the highest fixation densities throughout a video}. The highest \textit{density} of fixations is used in defining salient objects in video other than the highest \textit{number} of fixations. In this manner, we can avoid mistakenly assigning high saliency values to large background regions when salient objects are very small (\eg, the aircraft in Fig.~\ref{fig:badFixation} (b)).

\subsection{Generation of Salient Object Masks}
Based on the proposed definition, we can thus generate masks of salient objects for each video. We first compute the fixation density at each object in manually annotated keyframes. Considering that the fixations received by each keyframe are very sparse, we take the fixations recorded in a short period after the keyframe is displayed into consideration. Let $\mathcal{I}_t\in\mathcal{V}$ be a frame presented at time $t$ and $\mathcal{O}\in\mathcal{I}_t$ be an annotated object, we measure the fixation density at $\mathcal{O}$, denoted as $S_0(\mathcal{O})$, as
\begin{equation}\label{eq:fixationDensity}
\begin{split}
S_0(\mathcal{O})&=\frac{1}{\|\mathcal{O}\|}\sum_{f\in\mathbb{F}_\mathcal{V}}\delta(t_f>t)\\
&\cdot{}\left(\sum_{p\in{}\mathcal{O}}\text{Dist}(f,p)\cdot\exp\left(-\frac{(t_f-t)^2}{2\sigma_{t}^2}\right)\right),
\end{split}
\end{equation}
where $p$ is a pixel at $(x_p,y_p)$ and $\|\mathcal{O}\|$ is the number of pixels in $\mathcal{O}$. The indicator function $\delta(t_f>t)$ equals to 1 if $t_f>t$ and 0 otherwise. $\text{Dist}(f,p)$ measures the spatial distance between the fixation $f$ and the pixel $p$, which can be computed as
\begin{equation}\label{eq:spatialDist}
\text{Dist}(f,p)=\exp\left(-\frac{(x_f-x_p)^2+(y_f-y_p)^2}{2\sigma_{s}^2}\right).
\end{equation}
From \eqref{eq:fixationDensity} and \eqref{eq:spatialDist}, we can see that the influence of a fixation $f$ to the fixation density at the object $\mathcal{O}$ gradually decreases when the spatial or temporal distances between $f$ and pixels in $\mathcal{O}$ increase. Such influence is controlled by $\sigma_s$ and $\sigma_t$ which are empirically set to 3\% of video width (or video height if it is larger than the width) and 0.1s, respectively.

\begin{figure}[t]
\centering
\includegraphics[width=1.00\columnwidth]{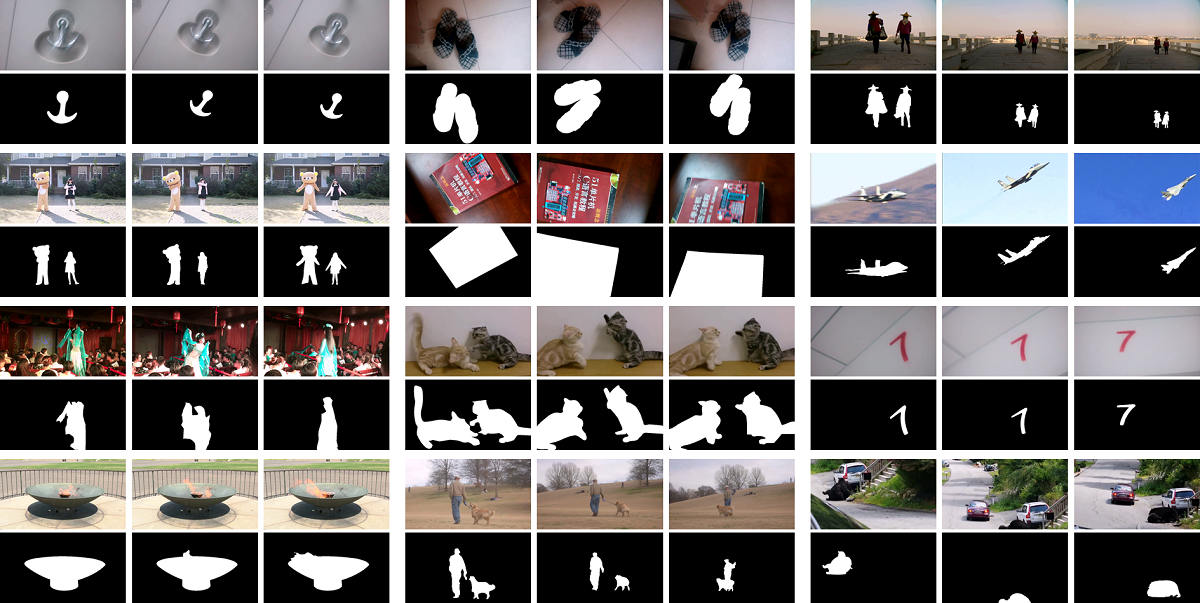}
\caption{Representative keyframes and masks of salient objects. }
\label{fig:exampleVOS}
\end{figure}

Based on the fixation density $S_0(\mathcal{O})$, we can thus compute its saliency score $S(\mathcal{O})$ from a global perspective:
\begin{equation}\label{eq:saliencyScore}
S(\mathcal{O})=\frac{\sum_{\mathcal{I}_t\in\mathcal{V}}\sum_{\mathcal{O}\in\mathcal{I}_t}S_0(\mathcal{O})}{\sum_{\mathcal{I}_t\in\mathcal{V}}\sum_{\mathcal{O}\in\mathcal{I}_t}1}.
\end{equation}
In \eqref{eq:saliencyScore}, the saliency of an object is defined as its average fixation density throughout a video. After that, we select the objects with saliency scores above an empirical threshold of 50 (or the object with the highest saliency score if it is smaller than 50). \rev{Note that such a threshold is empirically selected with respect to the (subjectively assessed) object completeness as well as the consistency between segmented salient objects and all recorded fixations}. Finally, we generate a set of salient objects for each video, represented by a sequence of binary masks at keyframes. In particular, a keyframe which contains only background or a salient object that splits into several disconnected parts due to the occlusion of background distractors will be discarded. Finally, we obtain $7,467$ binary masks of keyframes ($3,236$ for the 97 videos in \bl{VOS-E} and $4,231$ for the 103 videos in \bl{VOS-N}). Representative masks of salient objects can be found in Fig.~\ref{fig:exampleVOS}.

\begin{figure}[t]
\centering
\includegraphics[width=1.00\columnwidth]{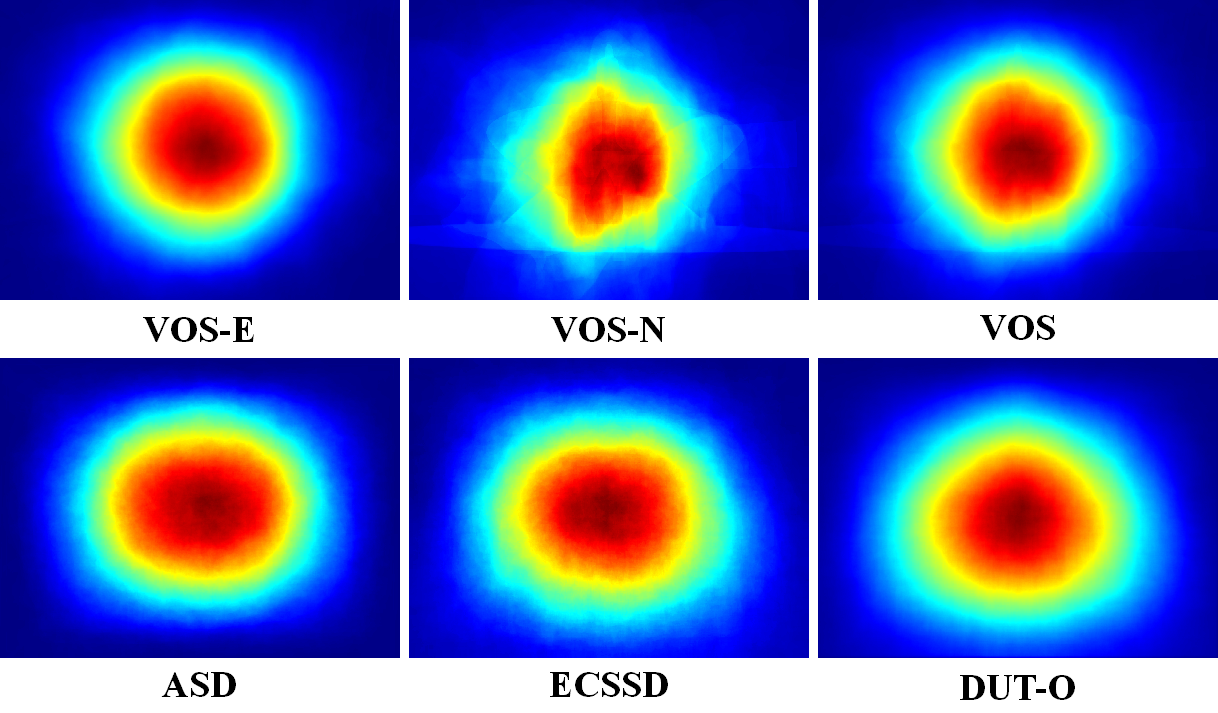}
\caption{The average annotation maps  of 6 datasets.}
\label{fig:averageMap}
\end{figure}
\subsection{Dataset Statistics}
To reveal the main characteristics of \bl{VOS}, we \rev{show} in Fig.~\ref{fig:averageMap} the average annotation maps (AAMs) of \bl{VOS-E}, \bl{VOS-N},  \bl{VOS} and three image datasets (\ie, \bl{ASD}~\cite{achanta2009frequency}, \bl{ECSSD}~\cite{yan2013hierarchical} and \bl{DUT-O}~\cite{YangZLRY13Manifold}). \rev{Similar to \cite{SalObjBenchmark}, the average annotation map (AAM) of an image-based SOD dataset is computed by 1)~resizing all ground-truth masks from the dataset into the same resolution, 2)~adding up the resized masks pixel by pixel, and 3)~normalizing the resulting map to a maximum value of 1.0. For a video-based SOD dataset (\eg, \bl{VOS-E}, \bl{VOS-N} and \bl{VOS}), an AAM is first computed over each video, while the AAMs from all videos are fused following the same three steps to obtain the final AAM. In this manner, we can provide a better view of the distribution of salient objects in different videos (otherwise the AAMs will be heavily influenced by long videos).}

From Fig.~\ref{fig:averageMap}, we can see that the distributions of salient objects in \bl{VOS} and its two subsets are both center-biased, while the degree of center-bias is a little stronger than that in \bl{ASD}, \bl{ECSSD} and \bl{DUT-O}. This is caused by the fact that photographers often have strong tendency to place salient targets near the center of the view in taking videos. This implies that image-based and video-based SOD are inherently correlated, and it is possible to directly transfer some useful saliency cues from the spatial domain to the spatiotemporal domain (\eg, the background prior~\cite{JiangWYWZL13,Zhang2015MB} obtained from the boundaries pixels).

\begin{figure}[t]
\centering
\includegraphics[width=1.00\columnwidth]{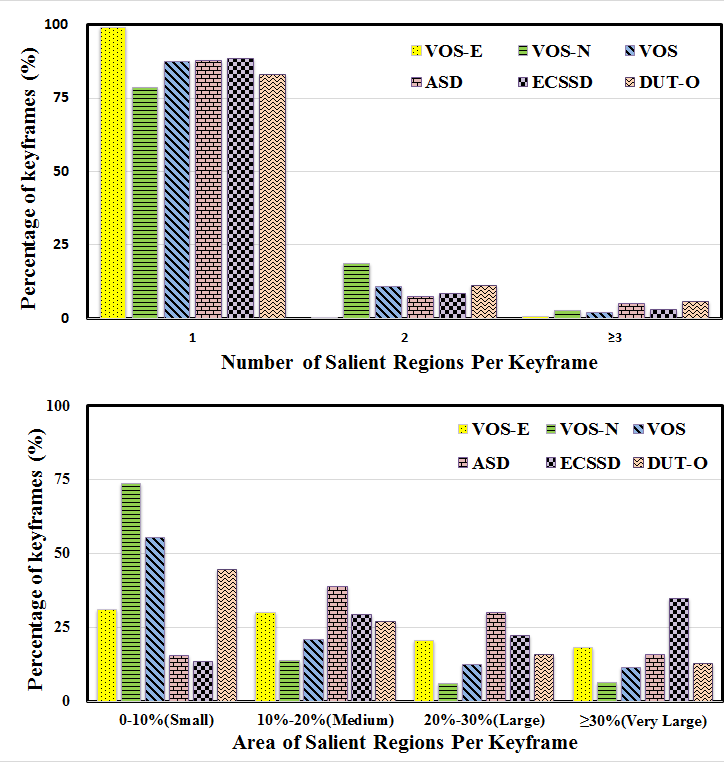}
\caption{\rev{Histograms of the number and area of salient objects.}}
\label{fig:objCount}
\end{figure}

Moreover, Figure~\ref{fig:objCount} shows the histograms of the number and area of salient objects. We see that the number and area of salient objects in \bl{VOS} are similar to those in \bl{DUT-O}. This implies that \bl{VOS}, like the \bl{DUT-O} dataset, is very challenging for reflecting many realistic scenarios. In particular, almost all keyframes from \bl{VOS-E} contain only one salient object, while the sizes of such salient objects distribute almost uniformly in the Small (31.1\%), Medium (30.1\%), Large (20.6\%) and Very Large (18.3\%) categories. This finding indicates that \bl{VOS-E} serves as a good baseline dataset to benchmark video-based SOD models.

% show the sum masks here.

\section{A Baseline Model for Video-based SOD with Saliency-guided Stacked Autoencoders}

\subsection{The Framework}

\begin{figure*}[t]
\centering
\includegraphics[width=1.00\textwidth]{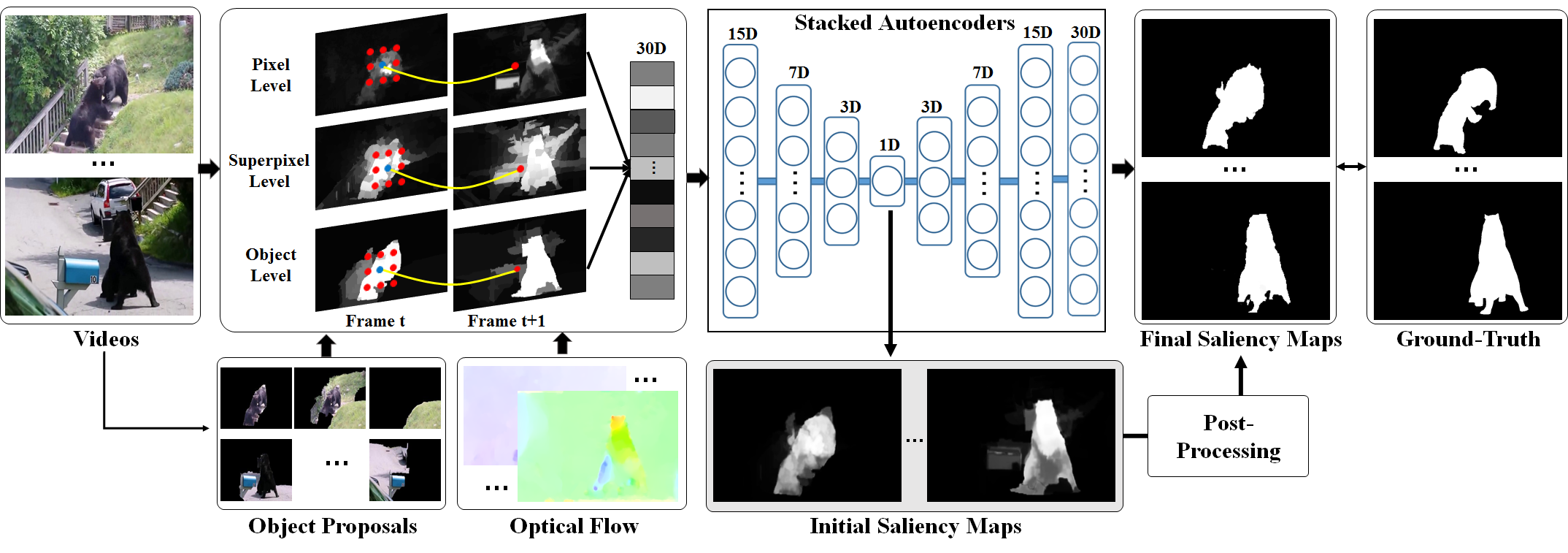}
\caption{The framework of the proposed saliency-guided stacked autoencoders.}
\label{fig:framework}
\end{figure*}

To construct a baseline model for \bl{VOS}, we propose an unsupervised approach that learns saliency-guided stacked autoencoders. The framework of the proposed approach is shown in Fig.~\ref{fig:framework}. We first turn each frame from \bl{VOS} into several color spaces and extract object proposals as well as the motion information (\eg, optical flow). After that, we extract three spatiotemporal saliency cues from each frame at pixel, superpixel and object levels, while such saliency cues reveal the presence of salient objects from different perspectives. Considering that salient objects are often spatially smooth and temporally consistent in consecutive frames, we characterize each pixel with a high-dimensional feature vector which consists of the saliency cues collected from the pixel, its spatial neighbors and the corresponding pixel in the subsequent frame.

With the guidance of saliency cues in the high dimensional feature vector at each pixel, stacked autoencoders can be unsupervisedly learned which contain only one hidden node in the last encoding layer (see Fig.~\ref{fig:framework}). Since the saliency cues within a pixel and its spatiotemporal neighbors can be well reconstructed from the output of this layer, we can safely assume that the degree of saliency at each pixel is strongly correlated with the output score. By computing the output scores and the linear correlation coefficient with the input saliency cues, we can derive an initial saliency map for each frame that is spatially smooth and temporally consistent. Finally, several simple post-processing operations are applied to further pop-out salient objects and suppress distractors.

\subsection{Extracting Multi-scale Saliency Cues}
To extract saliency cues, we first resize a frame $\mc{I}_t$ to the maximum side length of 400 pixels and convert it to the Lab and HSV color spaces. After that, we estimate the optical flow \cite{brox2010flow} between $\mc{I}_t$ and $\mc{I}_{t+1}$ and compute the inter-frame flicker as the absolute in-place difference of intensity between $\mc{I}_t$ and $\mc{I}_{t-1}$. For the sake of simplification, we use a space XYT formed by combining the optical flow and the flicker to indicate the variations along horizontal, vertical and temporal directions. Finally, each frame is represented by 12 feature channels from the RGB, Lab, HSV and XYT spaces. Based on these channels, we extract three types of saliency cues, including:

\myPara{1)~Pixel-based saliency}. To efficiently extract the pixel-based saliency, we refer to the algorithm proposed in \cite{Zhang2015MB} that computes the minimum barrier distance from a pixel to image boundary (one pixel width). In the computation, we discard the Hue channel since the substraction between hue values can not always
reflect the color contrast. Moreover, we also discard the RGB channels and the Value channel in HVS, which are somehow redundant to the other channels. For the rest 4 spatial and 3 temporal channels, the minimum barrier distances from all pixels to image boundary are separately computed over each channel. Such distances are then summed up across channels to initialize a pixel-based saliency map $\bl{S}_{t}^{pix}$. Moreover, we also extract a backgroundness map as in \cite{Zhang2015MB} and multiply it with $\bl{S}_{t}^{pix}$ to further enhance salient regions and suppress probable background regions. Finally, we conduct a morphological
smoothing step over the pixel-based saliency map to smooth $\bl{S}_{t}^{pix}$ while preserving the details of significant boundaries. As shown in Fig.~\ref{fig:saliencyCues} (c), the pixel-based saliency can be efficiently computed but sensitive to noise.

\myPara{2)~Superpixel-based saliency}. \rev{In image-based SOD, superpixels are often used as the basic units for feature extraction and saliency computation since they contain much
more structural information than pixels. In this study, we adopt the approach proposed in \cite{peng2016smd} to extract superpixel-based saliency in an unsupervised manner. This approach first divides a frame $\mc{I}_t$ into superpixels, base on which the sparse and low-rank properties are utilized to decompose the feature matrix of superpixels so as to obtain their saliency scores. In this process, prior knowledge on location (\ie, center-bias), color and background is used to refine the superpixel-based saliency. Finally, the saliency value of a superpixel is mapped back to all pixels it contains to generate a saliency map $\bl{S}_{t}^{sup}$. As shown in Fig.~\ref{fig:saliencyCues} (d), the superpixel-based saliency can detect a large salient object as a whole (\eg, the tissue in the third row of Fig.~\ref{fig:saliencyCues} (d)).}

\myPara{3)~Object-based saliency}. Inspired by the construction process of \bl{VOS}, we adopt the Multiscale Combinatorial Grouping algorithm \cite{pont2016mcg} to generate a set of object proposals for the frame $\mc{I}_t$ and estimate an objectness score for each proposal. After that, we adopt the unsupervised fixation prediction model proposed in \cite{li2013visual} to generate three fixation density maps in the Lab, HSV and XYT spaces, respectively. Let $\mb{O}$ be the top-ranked objects with the highest objectness scores and $\bl{F}_{lab},\bl{F}_{hsv},\bl{F}_{xyt}$ be the three fixation density maps, the object-based saliency at a pixel $p$ can be computed as:
\begin{equation}
\bl{S}_t^{obj}(p)=\sum_{\mc{O}\in{}\mb{O}}\delta(p\in{}\mc{O})\cdotp\bl{F}_{lab}(\mc{O})\cdot\bl{F}_{hsv}(\mc{O})\cdot\bl{F}_{xyt}(\mc{O}),
\end{equation}
where $\delta(p\in{}\mc{O})$ is an indicator function which equals to 1 if $p\in{}\mc{O}$ and 0 otherwise. $\mb{O}$ is the set of objects used for computing the object-based saliency maps, and we set $\|\mb{O}\|=50$ in experiments. $\bl{F}_{lab}(\mc{O})$ (or $\bl{F}_{hsv}(\mc{O}), \bl{F}_{xyt}(\mc{O})$) indicates the ratio of fixations received by $\mc{O}$ over the fixation density map $\bl{F}_{lab}$, which is computed as:
\begin{equation}
\bl{F}_{lab}(\mc{O})=\frac{\sum_{p\in\mc{O}}\bl{F}_{lab}(p)}{\sum_{p\in\mc{I}_t}\bl{F}_{lab}(p)}.
\end{equation}
As shown in Fig.~\ref{fig:saliencyCues} (e), the object-based saliency cues can successfully pop-out large salient object as a whole but often contain the background regions near them.

\subsection{Learning Stacked Autoencoders}
Given the saliency cues, we have to estimate a non-negative saliency score for each pixel, which, statistically, has positive correlation with the saliency cues.
Moreover, as stated in many previous works~\cite{papazoglou2013fst,wang2015GF,wang2015sag}, the estimated saliency scores should have the following attributes:

\myPara{1)~Spatial smoothness}. Similar pixels spatially adjacent to each other should have similar saliency scores.

\myPara{2)~Temporal consistency}. Corresponding pixels in adjacent frames should have similar saliency scores so that salient objects can consistently pop-out throughout a video.

To develop a model with such attributes, we train stacked autoencoders that take saliency cues at a pixel and its spatiotemporal neighbors as the input so that the spatial smoothness and temporal consistency of predicted saliency scores can be guaranteed. Considering the computational efficiency, for each pixel we adopt its eight spatial neighbors and only one temporal neighbor in the subsequent frame defined by the optical flow. A pixel is then represented by a feature vector with $3\times10=30$ saliency cues.

\begin{figure}[t]
\centering
\includegraphics[width=1.00\columnwidth]{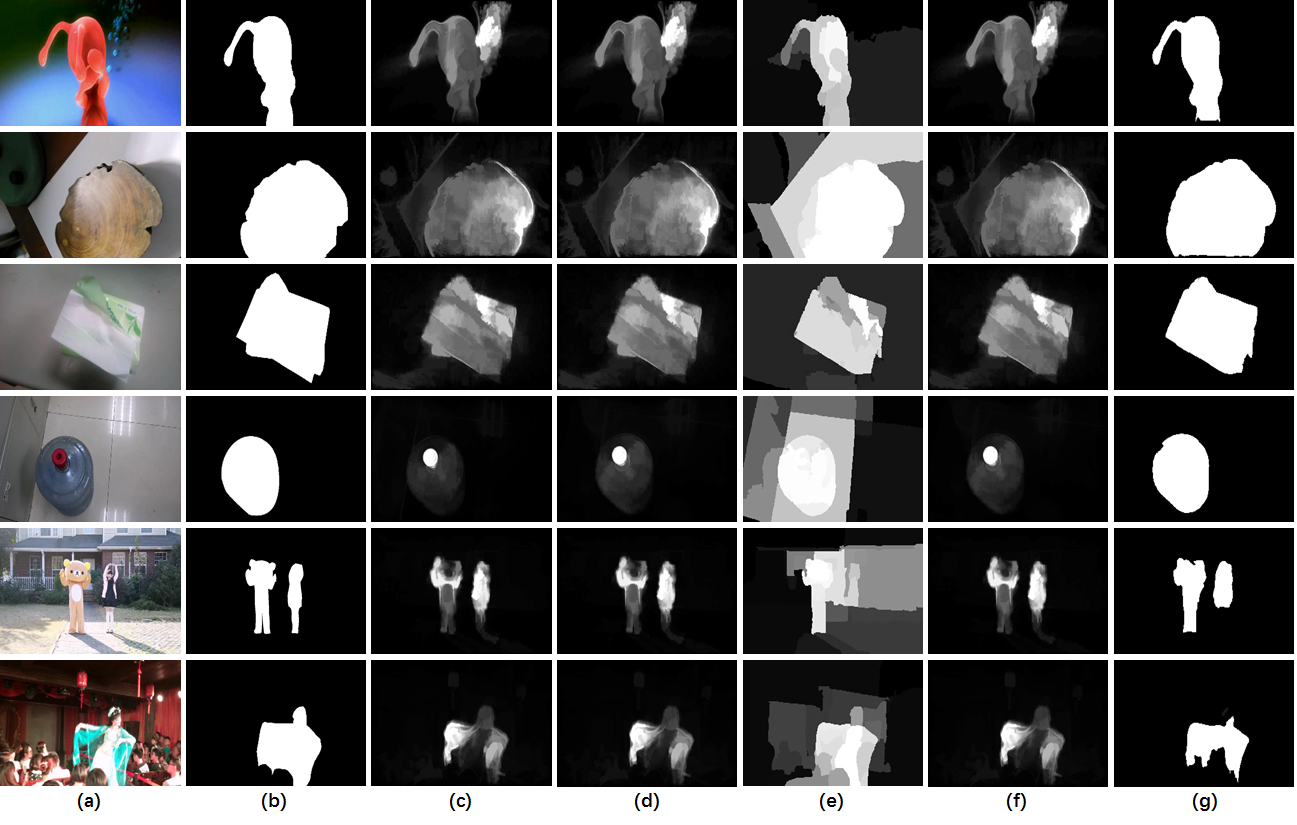}
\caption{Saliency cues and the estimated saliency maps. (a)~Frames, (b)~ground-truth, (c)~pixel-based saliency, (d)~superpixel-based saliency, (e)~object-based saliency, (f)~initial saliency maps obtained by the saliency-guided stacked autoencoders, (g)~final saliency maps obtained after post-processing.}
\label{fig:saliencyCues}
\end{figure}

With the guidance of the high-dimensional saliency cues, we collect the feature vectors from $N=500,000$ randomly selected pixels in \bl{VOS}, denoted as
$\{\bl{x}^1_n\}_{n=1}^{N}$. With these data, we train stacked autoencoders with $T$ encoding layers and the same number of decoding layers with logistic sigmoid transfer
functions. In the training process, no ground-truth data is used, while the $t$th encoding layer $f_t$, $t\in\{1,\ldots,T\}$ and its corresponding decoding layer $\hat{f}_t$ is trained by minimizing
\begin{equation}\label{eq:autoencoder}
\min_{f_t,\hat{f}_t}\frac{1}{N}\sum_{n=1}^N\|\bl{x}^t_n-\hat{f}_t\left(f_t(\bl{x}^t_n)\right)\|_2^2+\lambda_w\Omega_w+\lambda_s\Omega_s,
\end{equation}
where $\Omega_w$ is a $\ell$-2 regularization term that can be used to penalize the $\ell$-2 norm of weights in the encoding and decoding layers (we empirically set $\lambda_w=0.001$ in this study). $\Omega_s$ is a sparsity regularizer that is defined as the Kullback-Leibler divergence between the average output of each neuron in $f_t$ and a predefined score $\rho$ (we empirically set $\rho=0.05$ and $\lambda_s=1.0$).

In minimizing \eqref{eq:autoencoder}, the first encoding layer takes the sampled feature vectors of saliency cues as the input data, while other encoding layers take the output
of previous encoding layers as the input. That is, in training the $t$th encoding/decoding layer, we have
\begin{equation}
\bl{x}^t_n=\mc{N}\left(f_{t-1}(\bl{x}^{t-1}_n)\right),~\forall~t\in\{2,\ldots,T\},
\end{equation}
where $\mc{N}(\cdot)$ indicates the normalization operation that enforces each dimension of the input data that enters a encoding layer falls in the same dynamic range of $[-1,1]$. In this study, we use $T=4$ encoding layers with $15, 7, 3, 1$ neurons at each layer, and each layer is trained with 100 epochs. Note that the $T$th layer contains only one neuron, and by using its output score the input saliency cues within a pixel and its spatiotemporal neighbors can be well reconstructed by the decoding layers. As a result, we can safely assume that such output scores $\{\bl{x}^{T+1}_n\}_{n=1}^{N}$ are tightly correlated with the input saliency cues $\{\bl{x}^1_n\}_{n=1}^{N}$, and the the degree of correlation $c$ can be measured by averaging the linear correlation coefficients between $\{\bl{x}^{T+1}_n\}_{n=1}^{N}$ and every dimension of $\{\bl{x}^1_n\}_{n=1}^{N}$. As a result, the saliency score of a pixel $p$, given its feature vector $\bl{v}_p$ that contains the saliency cues from $p$ and its spatiotemporal neighbors, can be computed as
\begin{equation}\label{eq:salScore}
\bl{S}(p)=\text{sign}(c)\cdot{}f_T(\mc{N}(\cdots{}f_1(\mc{N}(\bl{v}_p)))).
\end{equation}
After computing a saliency score with \eqref{eq:salScore} for each pixel, we can initialize a saliency map for each frame in \bl{VOS} with the saliency values normalized to
$[0,255]$. As shown in Fig.~\ref{fig:saliencyCues} (f), such a saliency map already performs impressive in highlighting salient objects and suppressing distractors. To further pop-out salient objects and suppress distractors, we conduct three post-processing operations, including:
\begin{enumerate}
\item Apply temporal smoothing between adjacent frames to reduce the inter-frame flicker. We adopt a Gaussian filter with a width of 3 and $\sigma=0.75$.

\item Enhance the foreground/background contrast by using the sigmoid function proposed in \cite{Zhang2015MB}.

\item Binarize the saliency map with the average value of the whole saliency map and suppress the connected components that are extremely small.
\end{enumerate}
As shown in Fig.~\ref{fig:saliencyCues} (g),  these post-processing operations can generate compact and precise salient objects. Note that operations like center-biased re-weighting and spatial smoothing are not adopted here because the autoencoders unsupervisedly learned over a large-scale dataset already have the capability to accurately detect various types of salient objects despite their positions and sizes.

\section{Experiments}
In this Section, we compare the proposed Saliency-guided Stacked Autoencoders (SSA) with the state-of-the-art models on \bl{VOS}. The main objectives are two-fold: 1)~validate the effectiveness of the dataset \bl{VOS} and the baseline model \bl{SSA}, and 2)~provide a comprehensive benchmark to reveal the key challenges in video-based SOD. The rest of this section will first introduce the experimental settings and then discuss the results.

\begin{table}[t]
\scriptsize
\centering{
\caption{\rev{Models for benchmarking (Symbols: [I] for image-based, [V] for video-based; [C] for classic unsupervised or non-deep learning, [D] for deep learning, [U] for unsupervised).}}
\label{tab:models}
\begin{tabular}{r@{}r@{}r|r@{}r@{}r} \toprule
 ~ ~  Model~ ~                        & Pub.~\&~Year~\& &~Type   ~& Model~ ~                & Pub.~\&~Year~\& &~Type   \tabularnewline \midrule
\bl{SIV}~\cite{rahtu2010segmenting}     & ECCV 2010~&~~[V+U]   & \bl{CB}~\cite{jiang2011automatic}    & BMVC 2011~&~~[I+C]   \tabularnewline
\bl{RC}~\cite{ChengCVPR}                & CVPR 2011~&~~[I+C]   & \bl{ULR}~\cite{shen2012unified}      & CVPR 2012~&~~[I+C]   \tabularnewline
\bl{LMLC}~\cite{XieEtAlTIP2013}         & TIP  2013~&~~[I+C]   & \bl{DRFI}~\cite{JiangWYWZL13}        & CVPR 2013~&~~[I+C]   \tabularnewline
\bl{GMR}~\cite{YangZLRY13Manifold}      & CVPR 2013~&~~[I+C]   & \bl{HS}~\cite{yan2013hierarchical}   & CVPR 2013~&~~[I+C]   \tabularnewline
\bl{PCA}~\cite{margolinmakes}           & CVPR 2013~&~~[I+C]   & \bl{CHM}~\cite{LiLSDH13Contextual}   & ICCV 2013~&~~[I+C]   \tabularnewline
\bl{DSR}~\cite{li2013saliency}          & ICCV 2013~&~~[I+C]   & \bl{MC}~\cite{Jiang2013SaliencyMC}   & ICCV 2013~&~~[I+C]   \tabularnewline
\bl{FST}~\cite{papazoglou2013fst}       & ICCV 2013~&~~[V+U]   & \bl{HDCT}~\cite{kim2014salient}      & CVPR 2014~&~~[I+C]   \tabularnewline
\bl{RBD}~\cite{zhu2014saliency}         & CVPR 2014~&~~[I+C]   & \bl{NLC}~\cite{Faktor2014nlc}        & BMVC 2014~&~~[V+U]   \tabularnewline
\bl{BL}~\cite{Tong2015BL}               & CVPR 2015~&~~[I+C]   & \bl{BSCA}~\cite{Qin2015BSCA}         & CVPR 2015~&~~[I+C]   \tabularnewline
\bl{LEGS}~\cite{wang2015legs}           & CVPR 2015~&~~[I+D]   & \bl{MCDL}~\cite{zhao2015mcdl}        & CVPR 2015~&~~[I+D]   \tabularnewline
\bl{MDF}~\cite{li2015mdf}               & CVPR 2015~&~~[I+D]   & \bl{SAG}~\cite{wang2015sag}          & CVPR 2015~&~~[V+U]   \tabularnewline
\bl{GP}~\cite{Jiang2015GP}              & ICCV 2015~&~~[I+C]   & \bl{MB}~\cite{Zhang2015MB}           & ICCV 2015~&~~[I+C]   \tabularnewline
\bl{MB+}~\cite{Zhang2015MB}             & ICCV 2015~&~~[I+C]   & \bl{GF}~\cite{wang2015GF}            & TIP  2015~&~~[V+U]   \tabularnewline
\bl{ELD}~\cite{lee2016eld}              & CVPR 2016~&~~[I+D]   & \bl{DCL}~\cite{li2016dcl}            & CVPR 2016~&~~[I+D]   \tabularnewline
\bl{RFCN}~\cite{wang2016rfcn}           & ECCV 2016~&~~[I+D]   & \bl{DHSNet}~\cite{liu2016dhsnet}     & CVPR 2016~&~~[I+D]   \tabularnewline
\bl{SMD}~\cite{peng2016smd}             & PAMI 2017~&~~[I+C]   & \bl{SSA}                             & Our approach~&~~[V+U] \tabularnewline
\bottomrule
\end{tabular}
}
\end{table}
%\bl{ELE+}~\cite{xia2017what}            & CVPR 2017~&~[I+C]
%& \bl{ELE}~\cite{xia2017what}          & CVPR 2017~&~[I+C]   \tabularnewline

\begin{table*}[!t]
\centering{
\caption{Performance benchmarking of our approach and 31 state-of-the-art models on \bl{VOS} and its two subsets \bl{VOS-E} and \bl{VOS-N}. Top three scores in each column are marked in \rr{red}, \cg{green} and \bb{blue}, respectively. Symbols of model categories: [I+C] for image-based \& classic unsupervised or non-deep learning, [I+D] for image-based \& deep learning, [V+U] for video-based \& unsupervised.} \label{tab:performances}
\begin{tabular}{c@{}r|cccc|cccc|cccc} \toprule
\multicolumn{2}{c|}{\multirow{2}*{Models}} &  \multicolumn{4}{c|}{\bl{VOS-E}} & \multicolumn{4}{c|}{\bl{VOS-N}}     & \multicolumn{4}{c}{\bl{VOS}} \tabularnewline
                                  &  &~MAP~      &~MAR~     &~$F_\beta$~ &~MAE~   &~MAP~     & ~MAR~ &~$F_\beta$~ &~MAE~ &~MAP~   &~MAR~    &~$F_\beta$~ &~MAE~
\tabularnewline \midrule
\bl{\multirow{19}{*}{\rotatebox{90}{[I+C]}}}
&~\bl{CB}~\cite{jiang2011automatic}   & 0.755	   & 0.791    & 0.763    & 0.145    & 0.463	   & 0.563    & 0.483    & 0.229    & 0.605	   & 0.674    & 0.619    & 0.188    \tabularnewline
&~\bl{RC}~\cite{ChengCVPR}            & 0.738      & 0.677    & 0.723    & 0.171    & 0.465	   & 0.561    & 0.484    & 0.221    & 0.597	   & 0.617    & 0.602    & 0.197    \tabularnewline
&~\bl{ULR}~\cite{shen2012unified}     & 0.693	   & 0.737    & 0.703    & 0.158    & 0.390	   & 0.675    & 0.432    & 0.168    & 0.537	   & 0.705    & 0.568    & 0.163    \tabularnewline
&~\bl{LMLC}~\cite{XieEtAlTIP2013}     & 0.687	   & 0.736    & 0.697    & 0.154    & 0.408	   & 0.501    & 0.426    & 0.262    & 0.543	   & 0.615    & 0.558    & 0.210    \tabularnewline
&~\bl{GMR}~\cite{YangZLRY13Manifold}  & 0.813	   & 0.697    & 0.783    & 0.140    & 0.500	   & 0.611    & 0.522    & 0.195    & 0.652	   & 0.653    & 0.652    & 0.168    \tabularnewline
&~\bl{HS}~\cite{yan2013hierarchical}  & 0.755	   & 0.615    & 0.717    & 0.141    & 0.497	   & 0.521    & 0.502    & 0.262    & 0.622	   & 0.567    & 0.608    & 0.203    \tabularnewline
&~\bl{CHM}~\cite{LiLSDH13Contextual}  & 0.756	   & 0.765    & 0.758    & 0.124    & 0.409	   & 0.611    & 0.443    & 0.186    & 0.578    & 0.685    & 0.599	 & 0.156    \tabularnewline
&~\bl{DRFI}~\cite{JiangWYWZL13}       & 0.762	   & 0.837    & 0.778    & 0.114    & 0.442	   & 0.733    & 0.486    & 0.150    & 0.597	   & 0.783    & 0.632    & 0.132    \tabularnewline
&~\bl{PCA}~\cite{margolinmakes}       & 0.750	   & 0.725    & 0.744    & 0.143    & 0.420	   & 0.696    & 0.462    & 0.142    & 0.580	   & 0.710    & 0.606    & 0.143    \tabularnewline
&~\bl{DSR}~\cite{li2013saliency}      & 0.765	   & 0.748    & 0.761    & 0.112    & 0.450	   & 0.679	  & 0.488	 & 0.140    & 0.603	   & 0.713	  & 0.625	 & 0.127    \tabularnewline
&~\bl{MC}~\cite{Jiang2013SaliencyMC}  & 0.819      & 0.737    & 0.799    & 0.140    & 0.499	   & 0.665	  & 0.530    & 0.192    & 0.655	   & 0.700	  & 0.664	 & 0.167    \tabularnewline
&~\bl{HDCT}~\cite{kim2014salient}     & 0.711	   & 0.791    & 0.728    & 0.128    & 0.420    & 0.677    & 0.460    & 0.142    & 0.561	   & 0.733    & 0.593	 & 0.136    \tabularnewline
&~\bl{RBD}~\cite{zhu2014saliency}     & 0.799	   & 0.782    & 0.795    & 0.091    & 0.516    & 0.709    & 0.550    & 0.145    & 0.653	   & 0.745    & 0.672	 & 0.119    \tabularnewline
&~\bl{GP}~\cite{Jiang2015GP}          & 0.743	   & 0.788    & 0.753    & 0.141    & 0.405    & 0.704    & 0.449    & 0.227    & 0.569	   & 0.745	  & 0.602	 & 0.185    \tabularnewline
&~\bl{MB}~\cite{Zhang2015MB}          & 0.814	   & 0.735    & 0.794    & 0.107    & 0.480    & 0.696	  & 0.517    & 0.151    & 0.642	   & 0.715    & 0.657	 & 0.129    \tabularnewline
&~\bl{MB+}~\cite{Zhang2015MB}         & 0.803	   & 0.792    & 0.801    & 0.096    & 0.492    & 0.754    & 0.535    & 0.162    & 0.643	   & 0.772    & 0.669	 & 0.130    \tabularnewline
&~\bl{BL}~\cite{Tong2015BL}           & 0.765	   & 0.777    & 0.768    & 0.165    & 0.477    & 0.658    & 0.509    & 0.220    & 0.617    & 0.716    & 0.637	 & 0.194    \tabularnewline
&~\bl{BSCA}~\cite{Qin2015BSCA}        & 0.766	   & 0.758    & 0.764    & 0.133    & 0.457    & 0.663    & 0.493    & 0.195    & 0.607    & 0.709    & 0.628	 & 0.165    \tabularnewline
&~\bl{SMD}~\cite{peng2016smd}         & 0.811      & 0.789    & 0.806    & 0.096    & 0.528    & 0.688    & 0.558    & 0.148    & 0.665    & 0.737    & 0.681    & 0.123    \tabularnewline
\hline
%&~\bl{ELE}~\cite{xia2017what}         \tabularnewline
%&~\bl{ELE+}~\cite{xia2017what}        \tabularnewline
\bl{\multirow{7}{*}{\rotatebox{90}{[I+D]}}}
&~\bl{LEGS}~\cite{wang2015legs}       & 0.820       & 0.685    & 0.784    & 0.193    & 0.556    & 0.593    & 0.564    & 0.215    & 0.684    & 0.638    & 0.673    & 0.204    \tabularnewline
&~\bl{MCDL}~\cite{zhao2015mcdl}       & 0.831       & 0.787    & 0.821    & 0.081    & 0.570    & 0.645    & 0.586    & 0.085    & 0.697    & 0.714    & 0.701    & 0.083    \tabularnewline
&~\bl{MDF}~\cite{li2015mdf}           & 0.740       & 0.848    & 0.762    & 0.100    & 0.527    & 0.742    & 0.565    & 0.098    & 0.630    & 0.793    & 0.661    & 0.099    \tabularnewline
&~\bl{ELD}~\cite{lee2016eld}          & 0.790       &\bb{0.884}& 0.810    &\cg{0.060}& 0.569    &\cg{0.838}& 0.615    & 0.081    & 0.676    &\cg{0.861}& 0.712    &\cg{0.071}\tabularnewline
&~\bl{DCL}~\cite{li2016dcl}           &\cg{0.864}   & 0.735    & 0.830    & 0.084    & 0.583    &\bb{0.809}& 0.624    &\cg{0.079}& 0.719    & 0.773    & 0.731    & 0.081    \tabularnewline
&~\bl{RFCN}~\cite{wang2016rfcn}       & 0.834       & 0.820    &\bb{0.831}& 0.075    & 0.614    & 0.783    &\bb{0.646}&\bb{0.080}&\bb{0.721}&\bb{0.801}&\bb{0.738}&\bb{0.078}\tabularnewline
&~\bl{DHSNet}~\cite{liu2016dhsnet}    &\bb{0.863}   &\rr{0.905}&\rr{0.872}&\rr{0.049}&\cg{0.649}&\rr{0.851}&\rr{0.686}&\rr{0.055}&\cg{0.753}&\rr{0.877}&\rr{0.778}&\rr{0.052}\tabularnewline \hline
\bl{\multirow{6}{*}{\rotatebox{90}{[V+U]}}}
&~\bl{SIV}~\cite{rahtu2010segmenting} & 0.693	    & 0.543    & 0.651    & 0.204    & 0.451    & 0.523	   & 0.466    & 0.201    & 0.568	   & 0.533	  & 0.560    & 0.203    \tabularnewline
&~\bl{FST}~\cite{papazoglou2013fst}   & 0.781	    &\cg{0.903}& 0.806    & 0.076    &\bb{0.619}& 0.691	   & 0.634    & 0.117    & 0.697    & 0.794    & 0.718    & 0.097    \tabularnewline
&~\bl{NLC}$^*$~\cite{Faktor2014nlc}   & 0.439	    & 0.421	   & 0.435	  & 0.204    & 0.561	& 0.610	   & 0.572	  & 0.123    & 0.502	   & 0.518	  & 0.505	 & 0.162	\tabularnewline
&~\bl{SAG}~\cite{wang2015sag}         & 0.709	    & 0.814    & 0.731    & 0.129    & 0.354    & 0.742    & 0.402    & 0.150    & 0.526    & 0.777	  & 0.568    & 0.140    \tabularnewline
&~\bl{GF}~\cite{wang2015GF}           & 0.712       & 0.798    & 0.730    & 0.153    & 0.346    & 0.738    & 0.394    & 0.331   & 0.523    & 0.767   & 0.565    & 0.244    \tabularnewline
&~\bl{SSA}                            &\rr{0.875}   & 0.776    &\cg{0.850}&\bb{0.062}&\rr{0.660}& 0.682    &\cg{0.665}& 0.103   &\rr{0.764}& 0.728   &\cg{0.755}& 0.083    \tabularnewline
\bottomrule
\end{tabular}
\begin{tablenotes}
\footnotesize
\item[1] $^*$ The executable of NLC only output valid results on 187 videos (91 from \bl{VOS-E} and 96 from \bl{VOS-N}).
\end{tablenotes}
}
\end{table*}

% we add several new models at 2016.9.30, including MST (CVPR 2016),

\subsection{Settings}
As shown in Table \ref{tab:models}, thirty-two state-of-the-art models, including the proposed baseline model \bl{SSA}, are tested on the \bl{VOS} dataset (19 image-based \& classic unsupervised or non-deep learning, 7 image-based \& deep learning and 6 video-based \& unsupervised). Similar to many image-based SOD works, we also adopt Recall, Precision, $F_\beta$ and Mean Absolute Error (MAE) as the evaluation metrics. Let $G$ be the ground-truth binary mask of a keyframe and $S$ be the saliency map predicted by a model, the MAE score can be computed as the average absolute difference between all pixels in $S$ and $G$ to directly reflect the visual difference \cite{lee2016deep,SalObjBenchmark}. Moreover, the Recall and Precision scores can be computed by converting $S$ into a binary mask $M$ and comparing it with $G$:
\begin{equation}
\begin{split}
  \text{Recall}&=\frac{\#(\text{Non-zeros~in}~M\cap{}G)}{\#(\text{Non-zeros~in}~G)}, \\
  \text{Precision}&=\frac{\#(\text{Non-zeros~in}~M\cap{}G)}{\#(\text{Non-zeros~in}~M)},
\end{split}
\end{equation}
Intuitively, the overall performance of a model on \bl{VOS} can be assessed by directly computing the average Recall and Precision over all keyframes. However, this solution will over-emphasize the performance on long videos and ignore the performance on short videos (\eg, a video with 100 keyframes will overwhelm a video with only 10 keyframes). To avoid that, we first compute the average Recall, Precision and MAE separately over each video. After that, the mean values of the average Recall, Precision and MAE are computed over all videos. In this manner, the Mean Average Recall (MAR), Mean Average Precision (MAP) and MAE can well reflect the performance of a model by equivalently considering its performance over all videos. Correspondingly, $F_\beta$ is computed by fusing MAR and MAP to quantize the overall performance:
\begin{equation}
  F_\beta=\frac{(1+\beta^2)\text{MAP}\cdot{}\text{MAR}}{\beta^2\cdot\text{MAP}+\text{MAR}}.
\end{equation}
where we set $\beta^2=0.3$ as most of existing image-based models~\cite{achanta2009frequency,SalObjBenchmark} did in the performance evaluation.

Another problem in assessing models with MAP, MAR and $F_\beta$ is how to turn a gray-scale saliency map $S$ into a binary mask $M$. Similar to image-based SOD, we adopt the adaptive threshold proposed in \cite{achanta2009frequency}, which are computed as twice the average values of $S$, to generate a binary mask from each saliency map. Considering that such adaptive threshold may sometimes exceed the maximal saliency value of $S$ if there exists a very large salient object, we set this threshold to the maximal saliency value in this case. In this manner, unique MAR, MAP and $F_\beta$ scores can be generated to measure the overall performance of a model.

\begin{figure*}[t]
\centering
\includegraphics[width=0.96\textwidth]{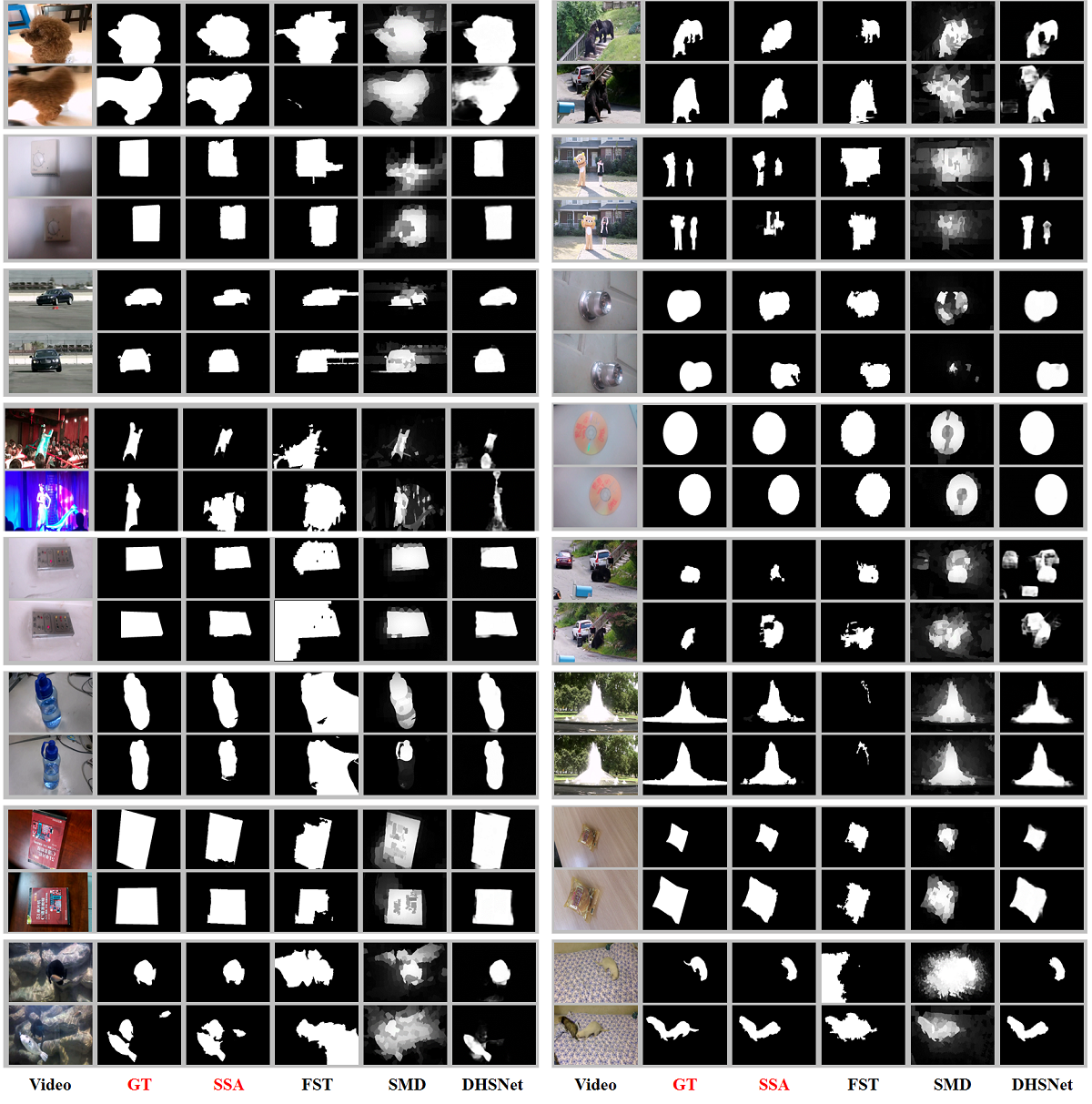}
\caption{The representative results of \bl{SSA} and the best model from each of the three model categories.}
\label{fig:results}
\end{figure*}

%\begin{figure*}[t]
%\centering
%\includegraphics[width=1.00\textwidth]{PRCurves.png}
%\caption{The Precision-Recall curves of our approach and 24 state-of-the-art models.}
%\label{fig:PRCurves}
%\end{figure*}

\subsection{Model Benchmarking}
The performance scores of the proposed baseline model \bl{SSA} and the other state-of-the-art models over \bl{VOS-E}, \bl{VOS-N} and \bl{VOS} are illustrated in Table~\ref{tab:performances}. Some representative results from the best model of each of the model categories are shown in Fig.~\ref{fig:results}. With Table \ref{tab:performances} and Fig.~\ref{fig:results}, we conduct several comparisons and discussions, including:

\myPara{1)~Comparisons between \bl{SSA} and the other models}. \rev{From Table~\ref{tab:performances}, we can see that \bl{SSA} outperforms 30 state-of-the-art models in terms of $F_\beta$, including 6 image-based deep models (except \bl{DHSNet}) and 5 video-based models. Note that no ground-truth data in any form has been used in \bl{SSA}, while the other deep models often make use of VGGNet~\cite{simonyan2014vgg16} pre-trained on massive images with semantic tags and have their SOD models fine-tuned on thousands of images with manually annotated salient objects (\eg, \bl{DHSNet} starts with VGGNet and then takes 9500 images from two datasets for model fine-tuning). Even in such an challenging setting, the unsupervised shallow model \bl{SSA}, which only utilizes four layers of stacked autoencoders, still outperforms all deep models in terms of MAP, and the $F_\beta$ score outperforms the other six deep models. This result validates the effectiveness of the saliency-guided autoencoding scheme in video-based SOD.}

\rev{One more thing that worth mentioning is that on \bl{VOS} and its two subsets, \bl{SSA} always has the best Precision ($\text{MAP}=0.764$ on \bl{VOS}), while its MAR scores are even lower than some unsupervised image-based models like \bl{MB+} and \bl{RBD}. This may be caused by the fact that such models adopt bottom-up frameworks that tend to pop-out almost all regions that are different from the predefined context (\ie, image boundary in \bl{MB+} and \bl{RBD}), leading to high recall rates. However, the suppression of distractors is less considered in such frameworks, making their precision much lower than \bl{SSA}. Actually, in the SOD task, it is widely recognized that a high precision is much more difficult to obtain than a high recall~\cite{liu2011learning,ChengPAMI}, and a frequently used trade-off is to gain a remarkable increase in precision at the cost of slightly decreasing recall.} That is why the computation of $F_\beta$ in this work and almost all the image-based models emphasize more on precision than recall. Although a higher recall usually leads to a better subjective impression in qualitative comparisons, the overall performance,  especially the $F_\beta$ score, may be not very satisfactory due to the emphasis of precision in computing $F_\beta$. This result also proposes a challenge for the proposed \bl{VOS} dataset: how to further improve the Recall rate while maintaining the high Precision?

\myPara{2)~Comparisons between (non-deep) image-based and video-based models}. Beyond analyzing the best models, another issue worth discussing is the performance of image-based and video-based models, especially the non-deep ones. Interestingly, video-based models like \bl{GF} and \bl{SAG} may sometimes perform even worse than the image-based models (\eg, \bl{SMD}, \bl{RBD} and \bl{MB+}). This may be caused by two reasons. First, the impact of incorporating temporal information in visual saliency computation is not always positive. In some videos, the salient objects, as assumed by many video-based models, have specific motion patterns that are remarkably different from the distractors (\eg, the dancing bear \& girl in the second row of Fig.~\ref{fig:exampleVOS}). However, such an assumption may not always hold in processing the ``realistic'' videos from \bl{VOS}. For example, in some videos with global camera motion and static salient objects/distractors (\eg, the shoes and book in the second column of Fig.~\ref{fig:exampleVOS}), the temporal information acts as a kind of noise and often leads to unsatisfactory results. Second, the parameters of most video-based models are manually fine-tuned on small datasets, which may become ``over-fitting'' to specific scenarios. Given a new scenario contained in \bl{VOS}, these parameters may lead to unsatisfactory results, either by emphasizing the wrong feature channels or by propagating the wrong results from some frames to the entire video in an energy-based optimization framework.

\myPara{\rev{3)~Comparisons between image-based deep and non-deep models}}. \rev{From Table~\ref{tab:performances}, we also find that image-based deep models often perform remarkably better than the image-based models with classic unsupervised or non-deep learning frameworks. This may be caused by the fact that deep models can be very complex to make use of massive training data. Taking the seven deep models compared in Table~\ref{tab:performances} as examples, we can find a ranked list with decreasing $F_\beta$ on \bl{VOS}. The ranked list, as well as their training data, are listed as follows: 1) \bl{DHSNet}: 9500 from \bl{MSRA10K} and \bl{DUT-O}, 2)~\bl{RFCN}: 10000 from \bl{MARA10K}, 3)~\bl{DCL}: 2500 from \bl{MSRA-B}, 4)~\bl{ELD}: 9000 from \bl{MSRA10K}, 5)~\bl{MCDL}: 8000 from \bl{MSRA10K}, 6)~\bl{LEGS}: 3340 from \bl{MSRA-B} and \bl{PASCAL-S}, and 7)~\bl{MDF}: 2500 from \bl{MSRA-B}. Note that the scenarios in \bl{DUT-O} and \bl{PASCAL-S} are much more challenging than those from \bl{MSRA-B} and \bl{MSRA10K} (many images of \bl{MSRA-B} are also contained in \bl{MSRA10K}). From this ranked list, we can conclude that, except an outlier \bl{DCL}, the more training data and training sources, the better performance of a deep model. This finding is quite interesting and may help to explain the success of some top-ranked deep models like \bl{DHSNet} and \bl{RFCN}. Moreover, the top-ranked models often adopt a recurrent mechanism in detecting salient objects, while such mechanisms can help to iteratively discover salient objects and suppress probable distractors. For video-based SOD, the success of such deep models shows a feasible way to develop better spatiotemporal models by using image-based training data as well as the recurrent architecture. Furthermore, it is necessary to develop an unsupervised baseline model that utilizes no training data in any form so as to provide fair comparisons for the other unsupervised and supervised models. That's why we propose \bl{SSA} that has the potential of being widely used as the baseline model on \bl{VOS}.}

\subsection{Performance Analysis of \bl{SSA}}

Beyond model benchmarking, we also conducted several experiments to analyze the performance of \bl{SSA}, including scalability test, influence of various components, influence of temporal window size, speed test and failure cases.

\myPara{\rev{1)~Scalability test}}. \rev{One concern about \bl{SSA} may be its scalability to other datasets. To validate this point, we reuse the stacked autoencoders generated on \bl{VOS} to a new dataset \bl{ViSal}~\cite{wang2015GF}. On \bl{ViSal}, the performance of \bl{SSA} and the other 9 models (\ie, the top three models on \bl{VOS} from each model categories) are reported in Table~\ref{tab:scalability}. We find that the overall performance of \bl{SSA}, although not fine-tuned on \bl{ViSal}, still ranks the second place on this dataset (only worse than the deep model \bl{DHSNet}). In particular, its MAE score even gains a higher rank than that on \bl{VOS}, which may be caused by the fact that \bl{VOS} is a large dataset that covers a variety of scenarios (\eg, \bl{VOS-N} contains many such outdoor scenarios about animal and aeroplane that also present in \bl{ViSal}). Moreover, the unsupervised architecture often have better performance in scalability test and can be generalized to new scenarios. This can be further proved by the model \bl{FST}, which ranks the third place in terms of $F_\beta$ on \bl{ViSal} (higher than its rank on \bl{VOS}). To sum up, \bl{VOS} contains a large number of real-world scenarios that may help to alleviate the over-fitting risk. Moreover, the unsupervised framework of \bl{SSA} makes it a scalable model that can be generalized to other scenarios without a remarkable performance drop.}

\begin{table}[!t]
\centering{
\caption{\rev{Performance scores of our approach and the other 9 models on \bl{ViSal}. The 9 models are selected as the top three models from each of the three model categories. The top 3 scores in each column are marked in \rr{red}, \cg{green} and \bb{blue}, respectively.} } \label{tab:scalability}
\begin{tabular}{c@{}r|cccc} \toprule
\multicolumn{2}{c|}{Models}                                &~MAP~        &~MAR~      &~$F_\beta$~ &~MAE~ \tabularnewline \midrule
\bl{\multirow{3}{*}{\rotatebox{90}{[I+C]}}}
&~\bl{MB+}~\cite{Zhang2015MB}         & 0.551       &\bb{0.887} & 0.604     & 0.145     \tabularnewline
&~\bl{RBD}~\cite{zhu2014saliency}     & 0.529       & 0.787     & 0.572     & 0.129  \tabularnewline
&~\bl{SMD}~\cite{peng2016smd}         & 0.583       & 0.886     & 0.633     & 0.133     \tabularnewline \hline
\bl{\multirow{3}{*}{\rotatebox{90}{[I+D]}}}
&~\bl{DCL}~\cite{li2016dcl}           & 0.718       & 0.859     & 0.747     & 0.261    \tabularnewline
&~\bl{RFCN}~\cite{wang2016rfcn}       & 0.781       &\cg{0.897} & 0.805     &\bb{0.050}    \tabularnewline
&~\bl{DHSNet}~\cite{liu2016dhsnet}    &\rr{0.816}   &\rr{0.955} &\rr{0.845} &\rr{0.027}\tabularnewline \hline
\bl{\multirow{4}{*}{\rotatebox{90}{[V+U]}}}
&~\bl{GF}~\cite{wang2015GF}           & 0.556       & 0.850     & 0.604     & 0.108    \tabularnewline
&~\bl{SAG}~\cite{wang2015sag}         & 0.538       & 0.858     & 0.589     & 0.104    \tabularnewline
&~\bl{FST}~\cite{papazoglou2013fst}   &\cg{0.803}   & 0.815     &\bb{0.806} & 0.052    \tabularnewline
&~\bl{SSA}                            &\bb{0.787}   & 0.884     &\cg{0.808} &\cg{0.046}\tabularnewline
\bottomrule
\end{tabular}
}
\end{table}

\myPara{\rev{2)~Influence of various components}}. \rev{\bl{SSA} involves three types of saliency, and we aim to explore which ones contribute the most to the performance of \bl{SSA}. Toward this end, we conduct an experiment to see the performance of \bl{SSA} on \bl{VOS} when some types of saliency are ignored. For fair comparisons, we adopt the same architecture of stacked autoencoders but replace some saliency cues to zeros in training and testing \bl{SSA}. As shown in Table~\ref{tab:componentInfluence}, the pixel-based saliency has the best precision, while object-based saliency has the best recall. Meanwhile, integrating all three types of saliency leads to the best overall performance. An interesting phenomena is that in the superpixel-only setting \bl{SSA} outperforms \bl{SMD} in both recall and precision, while \bl{SMD} is exactly the model used in computing the superpixel-based saliency. This may be mainly caused by the fact that temporal cues from adjacent frames are incorporated into the auto-encoding processes, which provides an opportunity to refine the results of \bl{SMD} from a temporal perspective. Due to the existence of temporal dimension in defining and annotating salient video objects, video-based SOD datasets contain something than cannot be obtained from image-based SOD datasets. For example, in the ``fighting bears'' scenario illustrated in the first two rows of the right column of Fig.~\ref{fig:results}, the mailbox and cars are considered to be non-salient from the perspective of the entire video, even though in some specific frames they do capture more human fixations than the fighting bears. In other words, the \bl{VOS} dataset provides a new way to explore the influence of spatiotemporal cues (\eg, optical flow and features propagated from adjacent frames) in defining, annotating and detecting salient objects, while in most image-based SOD datasets only spatial cues are involved. We believe the spatiotemporal definition of salient objects in \bl{VOS} may help future works to discover what is and what is not a salient object as the human-being does.
}

\begin{figure}[t]
\centering
\includegraphics[width=1.00\columnwidth]{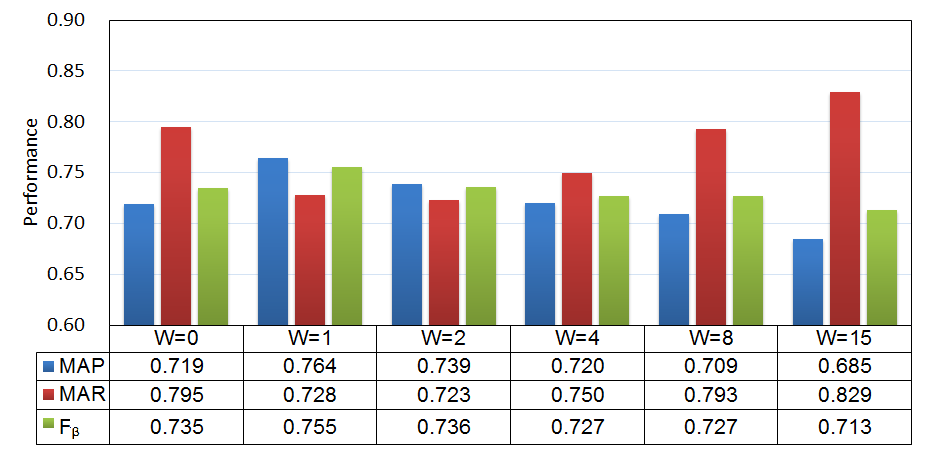}
\caption{Performance of \bl{SSA} on \bl{VOS} when temporal windows with different sizes are taken into consideration.}
\label{fig:influenceW}
\end{figure}

\myPara{\rev{3)~Influence of temporal window size}}. \rev{In \bl{SSA}, only one subsequent frame is referred to in processing a frame. To justify the rationality, we conduct an experiment that gradually incorporates none or more subsequent frames and show the $F_\beta$ variation of \bl{SSA} on \bl{VOS}. In this experiment, we refer to the next $W$ frames, while $W=0,1,2,4,8,15$. As shown in Fig.~\ref{fig:influenceW}, by referring only to the subsequent frame the $F_\beta$ score increase from 0.735 ($W$=0) to 0.755 ($W$=1). This result implies that the temporal cues can facilitate the detection of salient objects in a frame, even though consecutive frames are highly correlated. By incorporating more far-away frames at $W=2,4,8,15$, the performance gains are not as high as expected. This may be caused by the fact that the temporal correspondence between consecutive frames is the most reliable, while such reliability gradually decreases when the temporal gap between two frames increases. Such an experiment, together with the scalability test, can empirically prove that the over-fitting risk of \bl{SSA} is not very high, even though only one subsequent frame is used as the temporal context of the current frame.}

\myPara{\rev{4)~Speed test}}. \rev{The \bl{SSA} model consists of many feature extraction steps, and their speed analysis will help to find out how to further enhance the efficiency. Toward this end, we list the time costs of various key steps of \bl{SSA} in processing the first video in \bl{VOS}, and compare them with that of the other 5 video-based SOD models. Note that the video has original resolution $800\times448$, and we down-sample it to $400\times224$ for fair comparisons of various models in speed test. All models are tested on CPU platform (single core, 3.4GHz) with 128GB memory. As shown in Table~\ref{tab:speed}, the speed of \bl{SSA} is comparable to many previous algorithms like \bl{SIV} and \bl{NLC}. By investigating the time cost at each component of \bl{SSA}, we find that about 58.8\% computational resource is consumed in extracting object proposal. Moreover, about 22.1\% computational resource is spent on generating the optical flow. As a result, a probable way to speed up \bl{SSA} can be replacing these two components with faster models for optical flow computation and object proposal generation. In addition, the parallel processing mechanism can be explored as well, especially in extracting and encoding frame-wise saliency cues.}

\begin{table}[t]
%\scriptsize
\centering{
\caption{\rev{Performance of \bl{SSA} on \bl{VOS} when different types of saliency cues are used.}}
\label{tab:componentInfluence}
\begin{tabular}{ccccc}
\toprule
 Saliency Cues                 & MAP & MAR & $F_\beta$ & MAE  \tabularnewline
\midrule
Pixel-only                      & 0.705     & 0.649     & 0.691     & 0.105\tabularnewline
Superpixel-only                 & 0.691     & 0.744     & 0.703     & 0.103\tabularnewline
Object-only                     & 0.647     &\bl{0.811} & 0.678     & 0.155\tabularnewline
Pixel + Superpixel              & 0.677     & 0.801     & 0.702     & 0.100\tabularnewline
Pixel + Object                  &\bl{0.788} & 0.564     & 0.722     & 0.100\tabularnewline
Superpixel + Object             & 0.720     & 0.774     & 0.732     & 0.091\tabularnewline
All                             & 0.764     & 0.728     &\bl{0.755} &\bl{0.083}\tabularnewline
\bottomrule
\end{tabular}
}
\end{table}

\begin{table}[t]
%\scriptsize
\centering{
\caption{\rev{Speed test of \bl{SSA}, all its components and the other 5 video-based SOD models. All tests are tested on the first video of \bl{VOS} with 617 frames, which is down-sampled to the resolution of $400\times{}224$ for fair comparisons of all models.}}
\label{tab:speed}
\begin{tabular}{cc}
\toprule
 Models or Key Steps                  & Average Time (s/frame)  \tabularnewline
\midrule
\bl{SIV}~\cite{rahtu2010segmenting}   & 10.5 \tabularnewline
\bl{FST}~\cite{papazoglou2013fst}     & 5.80 \tabularnewline
\bl{NLC}~\cite{Faktor2014nlc}         &	19.0 \tabularnewline
\bl{SAG}~\cite{wang2015sag}           & 5.37 \tabularnewline
\bl{GF}~\cite{wang2015GF}             & 4.67  \tabularnewline
 \hline
Optical Flow                          & 1.84 \tabularnewline
Object proposal                       & 4.89 \tabularnewline
Pixel-based Saliency                  & 0.06 \tabularnewline
Superpixel-based Saliency             & 0.83 \tabularnewline
Object-based Saliency                 & 0.38 \tabularnewline
Auto-encoding \& Post-Proc.           & 0.31 \tabularnewline
\bl{SSA}                              & 8.31 \tabularnewline
\bottomrule
\end{tabular}
}
\end{table}

\myPara{5)~Failure cases}. Although \bl{SSA} achieves the best performance, we can see that its $F_\beta$ score is still far from perfect, which is mainly caused by the low Recall rate. On \bl{VOS-E} that contains only simple videos with nearly static salient objects and distractors as well as slow camera motion, \bl{SSA} only reaches a $F_\beta$ score of 0.850, while the performance score drops sharply to 0.665 on \bl{VOS-N}. This implies that the videos from the real-world scenarios are much more challenging than the videos taken in the laboratory environment. Actually, this is also the main reason that prevents the usage of existing SOD models in other applications.

\begin{figure}[t]
\centering
\includegraphics[width=1.00\columnwidth]{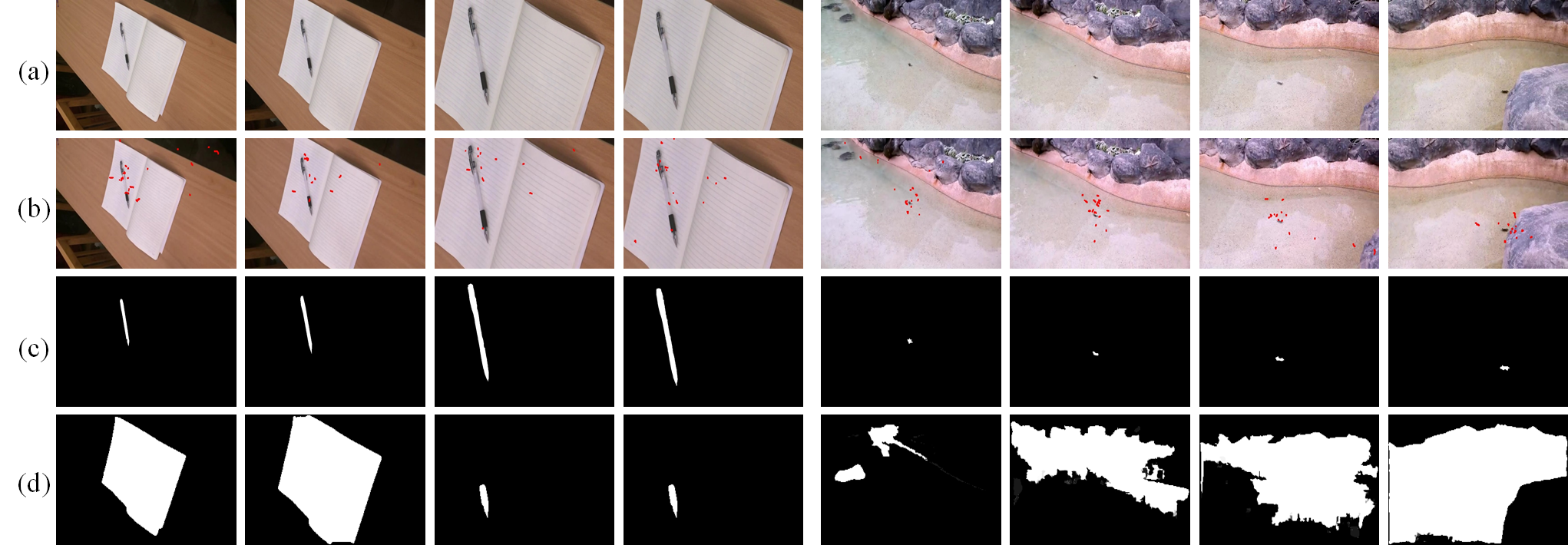}
\caption{Failure cases. (a)~Frames, (b)~the fixations received in 30ms after a keyframe is displayed, (c)~binary masks of salient objects and (d)~the estimated saliency maps of \bl{SSA}.}
\label{fig:failureCases}
\end{figure}

To validate this point, we illustrate in Fig.~\ref{fig:failureCases} two representative scenarios that \bl{SSA} fails, which actually provide two key challenges in video-based SOD. First, salient objects in a keyframe should be defined and detected by considering the entire video other than the keyframe itself. For example, in some early frames of Fig.~\ref{fig:failureCases} it is difficult to determine whether the pen or the notebook is the most salient object. Although in some later frames the pen is correctly detected, it is difficult to transfer such correct results to the frames far away. This indicates that the local spatiotemporal correspondences between pixels used by \bl{SSA} is still insufficient to handle more challenging scenarios, and a salient object should be detected by computing saliency from the global perspective as well.

Nevertheless, the failure cases in Fig.~\ref{fig:failureCases} not only suggest what should be considered in developing new video-based models but also validate the effectiveness of the \bl{VOS} dataset. Actually, the indoor/outdoor scenarios from \bl{VOS} are mainly taken by non-professional photographers, which are quite different from those in existing image datasets. For example, the moving crab in Fig.~\ref{fig:failureCases} consistently receives the highest density of fixations and becomes the most salient object in video, even though it is very small. The existence of such scenarios in \bl{VOS} increases the difficulties to transfer the knowledge learned on existing image datasets (\eg, the deep model \bl{DHSNet} learned from 9500 images) to the spatiotemporal domain, making video-based SOD on \bl{VOS} an extremely challenging task.  With such challenging cases, it is believed that \bl{VOS} can facilitate the development of new models by benchmarking their performance in processing real-world videos.

\subsection{Discussion}

From all the results presented above, we draw three major conclusions: First, video-based SOD is much more challenging than image-based SOD. Even the state-of-the-art image-based  models perform far from perfect without fully utilizing the temporal information from both local and global perspectives. Second, there exist some inherent correlations between image-based and video-based SOD, and the \bl{VOS-E} subset serves as a good baseline to help extend existing image-based models to the spatiotemporal domain. Third, real-world scenarios are still very challenging for existing models. In user-generated videos, salient objects may be very small, fast moving, with poor lighting conditions and cluttered dynamic background, etc. By handling such challenging scenarios in \bl{VOS-N}, a model can have better capability to process real-world scenarios. \rev{In particular, fixation prediction models often have impressive performance in detecting the most salient locations even in very complex real-world scenarios~\cite{han2016twostage,fang2014video}, developing a better fixation prediction model may be very helpful to handle the \bl{VOS-N} dataset in which salient objects are annotated with respect to human fixations.}

%table note: we use the optical flow computed by Liu Ce as an input of SIV.

\section{Conclusion}
Salient object Detection is a hot topic in the area of computer vision. In the past five years, hundreds of innovative models have been proposed for detecting salient objects in images, which gradually evolve from bottom-up models to deep models due to the presence of large-scale image datasets. However, the problem of video-based SOD has not been sufficiently explored since there lacks a large-scale video dataset. Actually, the most challenging part in building such a dataset is to provide a reasonable and unambiguous definition of salient objects from the spatiotemporal perspective.

To address this problem, this paper proposes \bl{VOS}, a large-scale dataset with 200 videos. Different from existing datasets, salient objects in \bl{VOS} are defined by combining human fixations and manually annotated objects throughout a video. As a result, the definition and annotation of salient objects in videos become less ambiguous. Moreover, we propose saliency-guided stacked autoencoders for video-based SOD, which, together with massive state-of-the-art models, are compared over \bl{VOS} to show the challenges of video-based SOD as well as its differences and correlations with image-based SOD. We find that \bl{VOS} is very challenging for containing a large amount of realistic videos, and its subset \bl{VOS-E} serves as a good baseline to extend existing image-based models to the spatiotemporal domain. Moreover, its subset \bl{VOS-N} covers many real-world scenarios that can help the deployment of better algorithms. This dataset can be very helpful for the area of video-based SOD, and the unsupervised saliency-guided stacked autoencoders can be used a good baseline model for benchmarking new video-based models.

%Q. Fan, F. Zhong, D. Lischinski, D. Cohen-Or, and B. Chen. JumpCut: Non-Successive Mask Transfer and Interpolation for Video Cutout. SIGGRAPH Asia, 2015. 2, 5, 6 （性能最好的方法）

%M. Grundmann, V. Kwatra, M. Han, and I. A. Essa. Efficient hierarchical graph-based video segmentation. In CVPR, 2010.

%A. Khoreva, F. Galasso, M. Hein, and B. Schiele. Classifier Based Graph Construction for Video Segmentation. In CVPR, 2015.

%Y. J. Lee, J. Kim, and K. Grauman. Key-segments for video object segmentation. In ICCV, 2011. 2, 7

%F. Li, T. Kim, A. Humayun, D. Tsai, and J. M. Rehg. Video segmentation by tracking many figure-ground segments. In ICCV, 2013. 4, 7

%saliency filter: .... find the code for comparison

%F. Perazzi, O. Wang, M. H. Gross, and A. Sorkine-Hornung. Fully connected object proposals for video segmentation. In ICCV, 2015. 2

%S. A. Ramakanth and R. V. Babu. Seamseg: Video object segmentation using patch seams. In CVPR, 2014. 5

%B. Taylor, V. Karasev, and S. Soatto. Causal Video Object Segmentation From Persistence of Occlusions. In CVPR, 2015. 2

%\section*{Acknowledgments}

%This work was supported in part by grants from the Chinese National Natural Science Foundation under contracts No. 61370113 and No. 61532003, and the Fundamental Research Funds for the Central Universities.

\bibliographystyle{IEEEtran}
\bibliography{egbib}

\begin{thebibliography}{10}
\providecommand{\url}[1]{#1}
\csname url@samestyle\endcsname
\providecommand{\newblock}{\relax}
\providecommand{\bibinfo}[2]{#2}
\providecommand{\BIBentrySTDinterwordspacing}{\spaceskip=0pt\relax}
\providecommand{\BIBentryALTinterwordstretchfactor}{4}
\providecommand{\BIBentryALTinterwordspacing}{\spaceskip=\fontdimen2\font plus
\BIBentryALTinterwordstretchfactor\fontdimen3\font minus
  \fontdimen4\font\relax}
\providecommand{\BIBforeignlanguage}[2]{{%
\expandafter\ifx\csname l@#1\endcsname\relax
\typeout{** WARNING: IEEEtran.bst: No hyphenation pattern has been}%
\typeout{** loaded for the language `#1'. Using the pattern for}%
\typeout{** the default language instead.}%
\else
\language=\csname l@#1\endcsname
\fi
#2}}
\providecommand{\BIBdecl}{\relax}
\BIBdecl

\bibitem{LiuSZTS07Learn}
T.~Liu, J.~Sun, N.~Zheng, X.~Tang, and H.-Y. Shum, ``Learning to detect a
  salient object,'' in \emph{IEEE Conference on Computer Vision and Pattern
  Recognition (CVPR)}, 2007, pp. 1--8.

\bibitem{achanta2009frequency}
R.~Achanta, S.~Hemami, F.~Estrada, and S.~S{\"u}sstrunk, ``Frequency-tuned
  salient region detection,'' in \emph{IEEE Conference on Computer Vision and
  Pattern Recognition (CVPR)}, 2009.

\bibitem{JiangWYWZL13}
H.~Jiang, J.~Wang, Z.~Yuan, Y.~Wu, N.~Zheng, and S.~Li, ``Salient object
  detection: A discriminative regional feature integration approach,'' in
  \emph{CVPR}, 2013, pp. 2083--2090.

\bibitem{Tong15bootstrap}
N.~Tong, H.~Lu, X.~Ruan, and M.-H. Yang, ``Salient object detection via
  bootstrap learning,'' in \emph{CVPR}, 2015, pp. 1884--1892.

\bibitem{han2016co}
D.~Zhang, D.~Meng, and J.~Han, ``Co-saliency detection via a self-paced
  multiple-instance learning framework,'' \emph{IEEE Transactions on Pattern
  Analysis and Machine Intelligence}, 2016.

\bibitem{Zhao15Deep}
R.~Zhao, W.~Ouyang, H.~Li, and X.~Wang, ``Saliency detection by multi-context
  deep learning,'' in \emph{CVPR}, 2015, pp. 1265--1274.

\bibitem{borji2012salient}
A.~Borji, D.~N. Sihite, and L.~Itti, ``Salient object detection: A benchmark,''
  in \emph{European Conference on Computer Vision (ECCV)}, 2012, pp. 414--429.

\bibitem{SalObjBenchmark}
A.~Borji, M.-M. Cheng, H.~Jiang, and J.~Li, ``Salient object detection: A
  benchmark,'' \emph{IEEE Transactions on Image Processing}, vol.~24, no.~12,
  pp. 5706--5722, 2015.

\bibitem{MSRA10KTHUR15Kdb}
MSRA10K and THUR15K, ``http://mmcheng.net/gsal/.''

\bibitem{YangZLRY13Manifold}
C.~Yang, L.~Zhang, H.~Lu, X.~Ruan, and M.-H. Yang, ``Saliency detection via
  graph-based manifold ranking,'' in \emph{IEEE Conference on Computer Vision
  and Pattern Recognition (CVPR)}, 2013.

\bibitem{yan2013hierarchical}
Q.~Yan, L.~Xu, J.~Shi, and J.~Jia, ``Hierarchical saliency detection,'' in
  \emph{IEEE Conference on Computer Vision and Pattern Recognition (CVPR)},
  2013.

\bibitem{borjiTIP2014}
A.~Borji, ``What is a salient object? a dataset and a baseline model for
  salient object detection,'' in \emph{IEEE Transactions on Image Processing},
  2014.

\bibitem{liXiaodiCVPR2014}
Y.~Li, X.~Hou, C.~Koch, J.~M. Rehg, and A.~L. Yuille, ``The secrets of salient
  object segmentation,'' in \emph{IEEE Conference on Computer Vision and
  Pattern Recognition (CVPR)}, 2014.

\bibitem{Peng14rgbd}
H.~Peng, B.~Li, W.~Xiong, W.~Hu, and R.~Ji, ``Rgbd salient object detection: a
  benchmark and algorithms,'' in \emph{ECCV}, 2014, pp. 92--109.

\bibitem{li2013co}
H.~Li, F.~Meng, and K.~Ngan, ``Co-salient object detection from multiple
  images,'' \emph{IEEE Transactions on Multimedia}, vol.~15, no.~8, pp.
  1896--1909, 2013.

\bibitem{Fu13Cluster}
H.~Fu, X.~Cao, and Z.~Tu, ``Cluster-based co-saliency detection,'' \emph{IEEE
  Transactions on Image Processing}, vol.~22, no.~10, pp. 3766--3778, 2013.

\bibitem{Zhang15MIL}
D.~Zhang, D.~Meng, C.~Li, L.~Jiang, Q.~Zhao, and J.~Han, ``A self-paced
  multiple-instance learning framework for co-saliency detection,'' in
  \emph{IEEE International Conference on Computer Vision (ICCV)}, 2015, pp.
  594--602.

\bibitem{rahtu2010segmenting}
E.~Rahtu, J.~Kannala, M.~Salo, and J.~Heikkil{\"a}, ``Segmenting salient
  objects from images and videos,'' in \emph{European Conference on Computer
  Vision (ECCV)}, 2010.

\bibitem{li2013exploring}
W.-T. Li, H.-S. Chang, K.-C. Lien, H.-T. Chang, and Y.-C.~F. Wang, ``Exploring
  visual and motion saliency for automatic video object extraction,''
  \emph{IEEE Transactions on Image Processing}, vol.~22, no.~7, pp. 2600--2610,
  2013.

\bibitem{Fu15Video}
H.~Fu, D.~Xu, B.~Zhang, S.~Lin, and R.~K. Ward, ``Object-based multiple
  foreground video co-segmentation via multi-state selection graph,''
  \emph{IEEE Transactions on Image Processing}, vol.~24, no.~11, pp.
  3415--3424, Nov 2015.

\bibitem{wang2015GF}
W.~Wang, J.~Shen, and L.~Shao, ``Consistent video saliency using local gradient
  flow optimization and global refinement,'' \emph{IEEE Transactions on Image
  Processing}, vol.~24, no.~11, pp. 4185--4196, Nov 2015.

\bibitem{tsai2010segtrack}
D.~Tsai, M.~Flagg, and J.~M.Rehg, ``Motion coherent tracking with multi-label
  mrf optimization,'' \emph{British Machine Vision Conference (BMVC)}, 2010.

\bibitem{perazzi2016davis}
F.~Perazzi, J.~Pont-Tuset, B.~McWilliams, L.~V. Gool, M.~Gross, and
  A.~Sorkine-Hornung, ``A benchmark dataset and evaluation methodology for
  video object segmentation,'' in \emph{IEEE Conference on Computer Vision and
  Pattern Recognition (CVPR)}, 2016.

\bibitem{brox2010longterm}
T.~Brox and J.~Malik, \emph{Object Segmentation by Long Term Analysis of Point
  Trajectories}.\hskip 1em plus 0.5em minus 0.4em\relax Berlin, Heidelberg:
  Springer Berlin Heidelberg, 2010, pp. 282--295.

\bibitem{li2015mdf}
G.~Li and Y.~Yu, ``Visual saliency based on multiscale deep features,'' in
  \emph{IEEE Conference on Computer Vision and Pattern Recognition (CVPR)},
  2015.

\bibitem{xia2017what}
C.~Xia, J.~Li, X.~Chen, A.~Zheng, and Y.~Zhang, ``What is and what is not a
  salient object? learning salient object detector by ensembling linear
  exemplar regressors,'' in \emph{IEEE Conference on Computer Vision and
  Pattern Recognition (CVPR)}, 2017.

\bibitem{li2013segtrackv2}
F.~Li, T.~Kim, A.~Humayun, D.~Tsai, and J.~M. Rehg, ``Video segmentation by
  tracking many figure-ground segments,'' in \emph{IEEE International
  Conference on Computer Vision (ICCV)}, 2013.

\bibitem{ochs2014moseg}
P.~Ochs, J.~Malik, and T.~Brox, ``Segmentation of moving objects by long term
  video analysis,'' \emph{IEEE Transactions on Pattern Analysis and Machine
  Intelligence}, vol.~36, no.~6, pp. 1187--1200, June 2014.

\bibitem{liu2011learning}
T.~Liu, Z.~Yuan, J.~Sun, J.~Wang, N.~Zheng, X.~Tang, and H.-Y. Shum, ``Learning
  to detect a salient object,'' \emph{IEEE Transactions on Pattern Analysis and
  Machine Intelligence}, vol.~33, no.~2, pp. 353--367, 2011.

\bibitem{li2009dataset}
J.~Li, Y.~Tian, T.~Huang, and W.~Gao, ``A dataset and evaluation methodology
  for visual saliency in video,'' in \emph{IEEE International Conference on
  Multimedia and Expo (ICME)}, 2009, pp. 442--445.

\bibitem{carmi2006role}
R.~Carmi and L.~Itti, ``The role of memory in guiding attention during natural
  vision,'' \emph{Journal of Vision}, vol.~6, no.~9, pp. 4, 898--914, 2006.

\bibitem{vigier2016uheddataset}
T.~Vigier, J.~Rousseau, M.~P. Da~Silva, and P.~Le~Callet, ``A new {HD} and
  {UHD} video eye tracking dataset,'' in \emph{International Conference on
  Multimedia Systems}, 2016, pp. 48:1--48:6.

\bibitem{ChengCVPR}
M.-M. Cheng, G.-X. Zhang, N.~J. Mitra, X.~Huang, and S.-M. Hu, ``Global
  contrast based salient region detection,'' in \emph{IEEE Conference on
  Computer Vision and Pattern Recognition (CVPR)}, 2011.

\bibitem{shen2012unified}
X.~Shen and Y.~Wu, ``A unified approach to salient object detection via low
  rank matrix recovery,'' in \emph{IEEE Conference on Computer Vision and
  Pattern Recognition (CVPR)}, 2012.

\bibitem{Jiang2013SaliencyMC}
B.~Jiang, L.~Zhang, H.~Lu, C.~Yang, and M.-H. Yang, ``Saliency detection via
  absorbing markov chain,'' in \emph{IEEE International Conference on Computer
  Vision (ICCV)}, 2013.

\bibitem{zhu2014saliency}
W.~Zhu, S.~Liang, Y.~Wei, and J.~Sun, ``Saliency optimization from robust
  background detection,'' in \emph{IEEE Conference on Computer Vision and
  Pattern Recognition (CVPR)}, 2014.

\bibitem{Zhang2015MB}
J.~Zhang, S.~Sclaroff, Z.~Lin, X.~Shen, B.~Price, and R.~Mech, ``Minimum
  barrier salient object detection at 80 fps,'' in \emph{IEEE International
  Conference on Computer Vision (ICCV)}, 2015, pp. 1404--1412.

\bibitem{yang2013manifold}
C.~Yang, L.~Zhang, H.~Lu, X.~Ruan, and M.-H. Yang, ``Saliency detection via
  graph-based manifold ranking,'' in \emph{IEEE Conference on Computer Vision
  and Pattern Recognition (CVPR)}, 2013.

\bibitem{ChengPAMI}
M.~M. Cheng, N.~J. Mitra, X.~Huang, P.~H.~S. Torr, and S.~M. Hu, ``Global
  contrast based salient region detection,'' \emph{IEEE Transactions on Pattern
  Analysis and Machine Intelligence}, vol.~37, no.~3, pp. 569--582, March 2015.

\bibitem{wang2015legs}
L.~Wang, H.~Lu, X.~Ruan, and M.-H. Yang, ``Deep networks for saliency detection
  via local estimation and global search,'' in \emph{IEEE Conference on
  Computer Vision and Pattern Recognition (CVPR)}, 2015.

\bibitem{li2016dcl}
G.~Li and Y.~Yu, ``Deep contrast learning for salient object detection,'' in
  \emph{IEEE Conference on Computer Vision and Pattern Recognition (CVPR)},
  2016.

\bibitem{zhao2015mcdl}
R.~Zhao, W.~Ouyang, H.~Li, and X.~Wang, ``Saliency detection by multi-context
  deep learning,'' in \emph{IEEE Conference on Computer Vision and Pattern
  Recognition (CVPR)}, 2015.

\bibitem{hou2017ehed}
Q.~Hou, M.-M. Cheng, X.-W. Hu, A.~Borji, Z.~Tu, and P.~Torr, ``Deeply
  supervised salient object detection with short connection,'' in \emph{IEEE
  Conference on Computer Vision and Pattern Recognition (CVPR)}, 2017.

\bibitem{han2015drr}
J.~Han, D.~Zhang, X.~Hu, L.~Guo, J.~Ren, and F.~Wu, ``Background prior-based
  salient object detection via deep reconstruction residual,'' \emph{IEEE
  Transactions on Circuits and Systems for Video Technology}, vol.~25, no.~8,
  pp. 1309--1321, 2015.

\bibitem{he2015supercnn}
S.~He, R.~Lau, W.~Liu, Z.~Huang, and Q.~Yang, ``Supercnn: A superpixelwise
  convolutional neural network for salient object detection,''
  \emph{International Journal of Computer Vision}, vol. 115, no.~3, pp.
  330--344, 2015.

\bibitem{lee2016eld}
G.~Lee, Y.-W. Tai, and J.~Kim, ``Deep saliency with encoded low level distance
  map and high level features,'' in \emph{IEEE Conference on Computer Vision
  and Pattern Recognition (CVPR)}, 2016.

\bibitem{srinivas2016su}
S.~S.~S. Kruthiventi, V.~Gudisa, J.~H. Dholakiya, and R.~V. Babu, ``Saliency
  unified: A deep architecture for simultaneous eye fixation prediction and
  salient object segmentation,'' in \emph{IEEE Conference on Computer Vision
  and Pattern Recognition (CVPR)}, 2016.

\bibitem{liu2016dhsnet}
N.~Liu and J.~Han, ``{DHSNet}: Deep hierarchical saliency network for salient
  object detection,'' in \emph{IEEE Conference on Computer Vision and Pattern
  Recognition (CVPR)}, 2016.

\bibitem{jason2016recurrent}
J.~Kuen, Z.~Wang, and G.~Wang, ``Recurrent attentional networks for saliency
  detection,'' in \emph{IEEE Conference on Computer Vision and Pattern
  Recognition (CVPR)}, 2016.

\bibitem{wang2016rfcn}
L.~Wang, L.~Wang, H.~Lu, P.~Zhang, and X.~Ruan, ``Saliency detection with
  recurrent fully convolutional networks,'' in \emph{European Conference on
  Computer Vision (ECCV)}, 2016.

\bibitem{liu2008video}
T.~Liu, N.~Zheng, W.~Ding, and Z.~Yuan, ``Video attention: Learning to detect a
  salient object sequence,'' in \emph{International Conference on Pattern
  Recognition (ICPR)}, 2008.

\bibitem{fukuchi2009saliencyseg}
K.~Fukuchi, K.~Miyazato, A.~Kimura, S.~Takagi, and J.~Yamato, ``Saliency-based
  video segmentation with graph cuts and sequentially updated priors,'' in
  \emph{IEEE International Conference on Multimedia and Expo (ICME)}, June
  2009, pp. 638--641.

\bibitem{bin2013temporally}
S.~Bin, Y.~Li, L.~Ma, W.~Wu, and Z.~Xie, ``Temporally coherent video saliency
  using regional dynamic contrast,'' \emph{IEEE Transactions on Circuits and
  Systems for Video Technology}, vol.~23, no.~12, pp. 2067--2076, 2013.

\bibitem{papazoglou2013fst}
A.~Papazoglou and V.~Ferrari, ``Fast object segmentation in unconstrained
  video,'' in \emph{IEEE International Conference on Computer Vision (ICCV)},
  Dec 2013, pp. 1777--1784.

\bibitem{wang2015sag}
W.~Wang, J.~Shen, and F.~Porikli, ``Saliency-aware geodesic video object
  segmentation,'' in \emph{IEEE Conference on Computer Vision and Pattern
  Recognition (CVPR)}, June 2015, pp. 3395--3402.

\bibitem{chiu2013coseg}
W.~C. Chiu and M.~Fritz, ``Multi-class video co-segmentation with a generative
  multi-video model,'' in \emph{IEEE Conference on Computer Vision and Pattern
  Recognition (CVPR)}, June 2013, pp. 321--328.

\bibitem{brox2010flow}
T.~Brox and J.~Malik, ``Large displacement optical flow: Descriptor matching in
  variational motion estimation,'' \emph{IEEE Transactions on Pattern Analysis
  and Machine Intelligence}, vol.~33, no.~3, pp. 500--513, March 2011.

\bibitem{peng2016smd}
H.~Peng, B.~Li, H.~Ling, W.~Hu, W.~Xiong, and S.~J. Maybank, ``Salient object
  detection via structured matrix decomposition,'' \emph{IEEE Transactions on
  Pattern Analysis and Machine Intelligence}, vol.~39, no.~4, pp. 818--832,
  April 2017.

\bibitem{pont2016mcg}
J.~Pont-Tuset, P.~Arbelaez, J.~Barron, F.~Marques, and J.~Malik, ``Multiscale
  combinatorial grouping for image segmentation and object proposal
  generation,'' \emph{IEEE Transactions on Pattern Analysis and Machine
  Intelligence}, 2016.

\bibitem{li2013visual}
J.~Li, M.~Levine, X.~An, X.~Xu, and H.~He, ``Visual saliency based on
  scale-space analysis in the frequency domain,'' \emph{IEEE Transactions on
  Pattern Analysis and Machine Intelligence}, vol.~35, no.~4, pp. 996--1010,
  2013.

\bibitem{jiang2011automatic}
H.~Jiang, J.~Wang, Z.~Yuan, T.~Liu, and N.~Zheng, ``Automatic salient object
  segmentation based on context and shape prior,'' in \emph{British Machine
  Vision Conference (BMVC)}, 2011.

\bibitem{XieEtAlTIP2013}
Y.~Xie, H.~Lu, and M.-H. Yang, ``Bayesian saliency via low and mid level
  cues,'' \emph{IEEE Transactions on Image Processing}, vol.~22, no.~5, 2013.

\bibitem{margolinmakes}
R.~Margolin, A.~Tal, and L.~Zelnik-Manor, ``What makes a patch distinct?'' in
  \emph{IEEE Conference on Computer Vision and Pattern Recognition (CVPR)},
  2013, pp. 1139--1146.

\bibitem{LiLSDH13Contextual}
X.~Li, Y.~Li, C.~Shen, A.~R. Dick, and A.~van~den Hengel, ``Contextual
  hypergraph modeling for salient object detection,'' in \emph{IEEE
  International Conference on Computer Vision (ICCV)}, 2013, pp. 3328--3335.

\bibitem{li2013saliency}
X.~Li, H.~Lu, L.~Zhang, X.~Ruan, and M.-H. Yang, ``Saliency detection via dense
  and sparse reconstruction,'' in \emph{IEEE International Conference on
  Computer Vision (ICCV)}, 2013.

\bibitem{kim2014salient}
J.~Kim, D.~Han, Y.-W. Tai, and J.~Kim, ``Salient region detection via
  high-dimensional color transform,'' in \emph{IEEE Conference on Computer
  Vision and Pattern Recognition (CVPR)}, 2014.

\bibitem{Faktor2014nlc}
A.~Faktor and M.~Irani, ``Video segmentation by non-local consensus voting,''
  in \emph{British Machine Vision Conference (BMVC)}, 2014.

\bibitem{Tong2015BL}
N.~Tong, H.~Lu, X.~Ruan, and M.-H. Yang, ``Salient object detection via
  bootstrap learning,'' in \emph{IEEE Conference on Computer Vision and Pattern
  Recognition (CVPR)}, 2015, pp. 1884--1892.

\bibitem{Qin2015BSCA}
Y.~Qin, H.~Lu, Y.~Xu, and H.~Wang, ``Saliency detection via cellular
  automata,'' in \emph{IEEE Conference on Computer Vision and Pattern
  Recognition (CVPR)}, 2015, pp. 110--119.

\bibitem{Jiang2015GP}
P.~Jiang, N.~Vasconcelos, and J.~Peng, ``Generic promotion of diffusion-based
  salient object detection,'' in \emph{IEEE International Conference on
  Computer Vision (ICCV)}, 2015, pp. 217--225.

\bibitem{lee2016deep}
G.~Lee, Y.~Tai, and J.~Kim, ``Deep saliency with encoded low level distance map
  and high level features,'' in \emph{IEEE Conference on Computer Vision and
  Pattern Recognition (CVPR)}, 2016.

\bibitem{simonyan2014vgg16}
K.~Simonyan and A.~Zisserman, ``Very deep convolutional networks for
  large-scale image recognition,'' \emph{CoRR}, vol. abs/1409.1556, 2014.

\bibitem{han2016twostage}
J.~Han, D.~Zhang, S.~Wen, L.~Guo, T.~Liu, and X.~Li, ``Two-stage learning to
  predict human eye fixations via {SDAEs},'' \emph{IEEE Transactions on
  Cybernetics}, vol.~46, no.~2, pp. 487--498, Feb 2016.

\bibitem{fang2014video}
Y.~Fang, W.~Lin, Z.~Chen, C.-M. Tsai, and C.-W. Lin, ``A video saliency
  detection model in compressed domain,'' \emph{IEEE Transactions on Circuits
  and Systems for Video Technology}, vol.~24, no.~1, pp. 27--38, 2014.

\end{thebibliography}

\end{document}